\documentclass{article}

\usepackage[preprint]{main}

\usepackage[utf8]{inputenc} 
\usepackage[T1]{fontenc}    
\usepackage{hyperref}       
\usepackage{url}            
\usepackage{booktabs}       
\usepackage{amsfonts}       
\usepackage{nicefrac}       
\usepackage{microtype}      
\usepackage{xcolor}         

\usepackage{graphicx}
\usepackage{subfigure}

\usepackage{amsmath}
\usepackage{amssymb}
\usepackage{mathtools}
\usepackage{amsthm}
\usepackage{multirow}
\usepackage[capitalize,noabbrev]{cleveref}

\theoremstyle{plain}
\newtheorem{theorem}{Theorem}[section]

\newtheorem{lemma}[theorem]{Lemma}

\theoremstyle{definition}

\theoremstyle{remark}

\usepackage{algorithm}
\usepackage{algorithmic}

\usepackage{wrapfig}

\title{Collaborative AI Teaming in Unknown Environments via Active Goal Deduction}

\author{
    Zuyuan Zhang\\
    The George Washington University\\
    \texttt{zuyuan.zhang@gwu.edu}\\
    \And
    Hanhan Zhou\\
    The George Washington University\\
    \texttt{hanhan@gwu.edu}\\
    \And
    Mahdi Imani\\
    Northeastern University\\
    \texttt{m.imani@northeastern.edu}
    \And
    Taeyoung Lee\\
    The George Washington University\\
    \texttt{tylee@gwu.edu}\\
    \And
    Tian Lan\\
    The George Washington University\\
    \texttt{tlan@gwu.edu}
}

\begin{document}

\maketitle

\begin{abstract}

With the advancements of artificial intelligence (AI), we're seeing more scenarios that require AI to work closely with other agents, whose goals and strategies might not be known beforehand. 
However, existing approaches for training collaborative agents often require defined and known reward signals and cannot address the problem of teaming with unknown agents that often have latent objectives/rewards.
In response to this challenge, we propose teaming with unknown agents framework, which leverages kernel density Bayesian inverse learning method for active goal deduction and utilizes pre-trained, goal-conditioned policies to enable zero-shot policy adaptation. 
We prove that unbiased reward estimates in our framework are sufficient for optimal teaming with unknown agents.
We further evaluate the framework of redesigned multi-agent particle and StarCraft II micromanagement environments with diverse unknown agents of different behaviors/rewards. Empirical results demonstrate that our framework significantly advances the teaming performance of AI and unknown agents in a wide range of collaborative scenarios.
\end{abstract}

\section{Introduction}
\label{submission}
Advancements in machine learning and artificial intelligence (ML/AI) are enabling more and more scenarios where AI agents are to collaborate with other autonomous systems or humans, which are often considered unfamiliar entities outside the environments with unknown objectives~\cite{johnson2020understanding,cooperative_AI,tao2022optimal}.
Examples include teaming with
autonomous agents that were built by other developers with unknown designs/parameters~\cite{traeger2020vulnerable,albrecht2018autonomous}, or humans in a shared work environment with undefined or only partially defined intents/goals~\cite{simmler2021taxonomy,behymer2016autonomous}. The ability to team up with such unknown agents and to effectively collaborate toward common (yet often latent) objectives can be crucial for solving complex tasks that would be otherwise impossible~\cite{cooperative_AI}. Existing methods of training AI agents, such as multi-agent reinforcement learning (MARL)~\cite{POMDP} and transfer learning~\cite{weiss2016survey,yu2021decentralized}, often cannot support synergistic teaming with unknown agents, due to the absence of pre-defined goals and rewards. 

In this paper, we propose a novel framework to develop AI agents for \textbf{S}ynergistic \textbf{T}eaming with \textbf{UN}known agents (STUN), through active goal inference and zero-shot policy adaptation. Specifically, in collaborative task environments, we leverage inverse learning to enable AI agents to actively reason and infer the reward signals (i.e., the posterior distribution) of the unknown agents from their observed trajectories on the fly. 
Then, we show that obtaining an unbiased reward estimate is sufficient to ensure the optimality of learning collaborative policies. Based on this result, we utilize the inferred reward signal to achieve a zero-shot policy adaptation by pre-training collaborative AI agent policies with respect to randomly sampled surrogate models.
This novel teaming framework goes beyond existing approaches, which either fail to operate in the absence of reward signals (e.g., policy re-training and transfer learning~\cite{weiss2016survey}) or resort to general one-size-fits-all policies with non-optimal teaming performance (e.g., multi-task learning~\cite{zhang2018overview} with respect to assumed unknown reward distributions).  

\begin{wrapfigure}{r}{0.5 \columnwidth}
\vspace{-0.1in}
\begin{center}
\centerline{\includegraphics[width=0.5 \columnwidth]{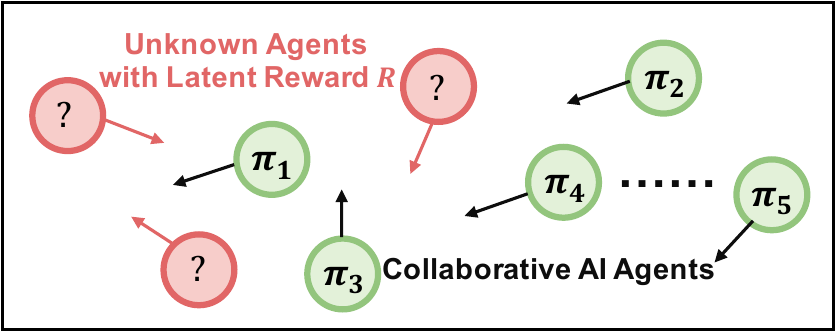}}
\vspace{-0.12in}
\caption{\footnotesize We consider the problem of enabling synergistic teaming of AI agents with other unknown agents (e.g., human or autonomous agents that could have latent rewards/objectives) in collaborative task environments. }
\vspace{-0.4in}
\label{fig:illu}
\end{center}
\end{wrapfigure}

For active goal inference, we propose a Kernel Density Bayesian Inverse Learning (KD-BIL) to obtain a posterior estimate of the latent goal/reward function of the unknown agents. This method is sample efficient and eliminates the need to refit policies for each sampled reward function by utilizing the kernel density estimation to approximate the likelihood function. The kernel density function represents the probability of observing certain states and actions given a reward function~\cite{KD_BIRL}. It allows efficient posterior inference of reward functions in light of observed sequences from agents.

Interestingly, directly using the maximum a posteriori (MAP) of latent rewards cannot ensure the optimality of learning collaborative policies. To ensure convergence and optimality of the Bellman equation, we prove that obtaining an unbiased estimation is necessary. This result motivates us to develop a zero-shot policy adaptation strategy for teaming with unknown agents in STUN. It leverages a decentralized partially observable Markov Decision Process (dec-POMDP) model~\cite{DEC_POMDP} to pre-train collaborative agent policies with respect to randomly sampled surrogate models (of the unknown agents), such that the learned collaborative policies $\pi_i(a_i|s_i,R)$ (known as the goal-conditioned policies) are conditioned on potential rewards $R$. To team up with unknown agents, a zero-shot policy adaptation can be easily achieved by conditioning the learned collaborative policies on the unbiased estimation $\hat{R}$ (obtained by active goal inference from observed unknown agents' trajectories), i.e., $\pi_i(a_i|s_i,\hat{R})$. The proposed STUN framework is scalable as it is based on centralized pre-training and fully decentralized executions~\cite{kraemer2016multi}.

To validate the effectiveness of the STUN framework, we created the first coop environments for teaming and collaborating with unknown agents by modifying the MPE and SMAC environments~\cite{MPE,SMAC}. We redesigned the reward system to reflect various latent play styles.  The blue team is then composed of both collaborative AI agents and unknown agents (e.g., downloaded from public repositories or trained with latent rewards using popular MARL algorithms like MAPPO, IPPO, COMA, and IA2C ~\cite{MAPPO,IPPO,COMA,IA2C}). Compared with a wide range of baselines, STUN agents consistently achieve close-to-optimal teaming performance with unknown agents in almost all scenarios/tasks. On super hard maps like $27m\_vs\_30m$ or $mmm2$, it improves the reward of unknown agents by up to 50\%. It also demonstrates the ability to cognitively reason and adapt to the non-stationarity of unknown agents.

\section{Related Work}

\noindent {\bf Human AI teaming:}
Existing work on human-AI teaming often focuses on team dynamics and organizational behavior contributes to understanding how to build effective human-AI teams~\cite{cooperative_AI,albrecht2018autonomous}, leveraged cognitive science to better model and complement human decision-making processes~\cite{hu2023language,traeger2020vulnerable}, and considered related issues such as communication, trust, and collaboration strategies~\cite{bauer2023human}. Other related work is as follows~\cite{chen2024deep}. However, modeling humans in a shared task environment as unknown agents and supporting human-AI teaming through active goal inference and POMDP models have not been considered.

\noindent {\bf Multi-agent Reinforcement Learning: } Most of the successful RL applications, e.g., gaming~\cite{iqbal2019actor,foerster2017stabilising,mei2024projection,mei2023mac} and robotics~\cite{knapp2013nonverbal,robinette2016overtrust}, involve the participation of multiple agents. For collaborative agents~\cite{matignon2007hysteretic,panait2006lenient}, MARL learns joint decision-making policies to optimize a shared reward. These problems are often formulated as decentralized POMDPs and solved using policy- or value-based methods like MAPPO~\cite{MAPPO}, MATRPO~\cite{MATRPO}, and factorization~\cite{WQMIX,PAC}. Existing work has also considered transfer learning in this context~\cite{yang2021efficient}. Other related work is as follows~\cite{zhou2023value,zhou2023double,chen2021bringing,chen2023distributional}.
However, teaming and collaborating with unknown agents with undefined rewards are underexplored.

\noindent {\bf Multi-task Learning:}  
Multi-task learning aims to train intelligence to solve multiple tasks simultaneously~\cite{multi_task_1}. 
Negative Transfer (NT) between tasks is a major challenge in multi-task RL, i.e., knowledge learned while solving one task may interfere with the learning of other tasks \cite{rusu2022progressive,omidshafiei2017deep,gurulingan2023multi}. Methods like Modular Neural Networks (MNNs)~\cite{auda1999modular} and Attention Mechanisms (AMs) ~\cite{niu2021review} are proposed to reduce negative transfer \cite{feudal,wang2023improved}. Other related work is as follows~\cite{chen2023real}. Multi-task learning can produce general policies that work in different task environments but may not achieve optimal teaming performance with specific unknown agents.

\noindent {\bf Inverse Reinforcement Learning:} Inverse Reinforcement Learning (IRL) infers goals/rewards from observations of action trajectories. It was first proposed in~\cite{ng2000algorithms} showing that the IRL problem has infinite solutions. Several solutions, such as MaxEntropy IRL~\cite{ziebart2008maximum}, Max-Margin IRL~\cite{abbeel2004apprenticeship}, and Bayesian IRL~\cite{ramachandran2007bayesian} have been proposed to resolve the ambiguity of IRL. Other related work is as follows~\cite{mei2022bayesian,mei2024bayesian}. In contrast, kernel-based IRL in this paper is more sample-efficient and supports synergistic teaming with unknown agents on the fly.

\section{Our Proposed Solution}
\subsection{Problem Statement}
Consider a 
dec-POMDP model~\cite{DEC_POMDP} involving both unknown agents and collaborative AI agents (also denoted as STUN agents), given by a tuple  $\mathcal{M} =\langle \mathcal{N}_u, \mathcal{N}, \mathcal{S}, \mathcal{A},\mathcal{T}, R ,\Omega, \mathcal{O}, \gamma  \rangle$. $\mathcal{N}_u$ denotes a set of $n_u$ unknown agents with a latent reward $R$, while $\mathcal{N}$ denotes a set of $n$ AI-agents supporting the unknown agents toward the latent goal. $\mathcal{S}$ is the joint state space for all agents $\mathcal{N}_u\cup \mathcal{N}$. For each agent $i$, $A_{i}$ is its action space and $O_{i}$ its observation space. Thus, $\mathcal{A} = A_{1}\times A_{2}\times \dots \times A_{n_u+ n}$ denotes the joint action space of all agents, and $\mathcal{O} = O_{1}\times O_{2}\times \dots \times O_{n_u+ n}$ denotes the joint observation space of all agents. We use $\mathcal{T}:\mathcal{S} \times \mathcal{A}\rightarrow \Omega(\mathcal{S})$ to denote the transition probability function, with $\Omega(\mathcal{S})$ representing a set of state distributions. 

Each unknown agent $i\in \mathcal{N}_u$ behaves according to some latent policy $\pi^u_{i}:O_{i}\times A_{i} \rightarrow [0,1]$, which is a probability distribution representing the probability of the agent taking each action in $A_{i}$ under observation $O_{i}$. We assume that the unknown agents are logical decision-makers -- their behaviors $\pi^u$ are aligned with and maximize the latent reward $R$. The latent reward function $R: \mathcal{S}\times A \rightarrow \mathbb{R}$ is unknown and non-observable to
other AI agents operating in the shared task environment, while their behavior trajectories are observable for reasoning and inferring the latent reward (which could be time-varying).  

Our goal in this paper is to learn the policies of collaborative AI agents (i.e., STUN agents):  $\pi_{i}:O_{i}\times A_{i} \rightarrow [0,1]$ for $i\in \mathcal{N}$, to effectively team up with the unknown agents and collaboratively maximize the expected return $G = \sum^{H}_{t}\gamma^{t}R^{t}$, where $\gamma$ is a discount factor, $R^t$ is the latent reward received at time $t$,  and $H$ is the time horizon. It is easy to see that while the problem follows a dec-POMDP structure, it cannot be solved with existing MARL algorithms because training the MARL agents would require having access to the latent reward signal $R$, and thus is not possible in tasks/scenarios teaming up with unknown agents. The collaborative AI agents must simultaneously address two problems: (i) inferring the latent reward by collecting observations of the unknown agents in the shared task environment and (ii) adapting their policies on the fly to support effective teamwork toward the learned reward without incurring significant overhead such as re-training.

\subsection{Overview of Our STUN framework}
\begin{wrapfigure}{r}{0.5 \columnwidth}
\vspace{-0.1in}
\begin{center}
\centerline{\includegraphics[width=0.5 \columnwidth]{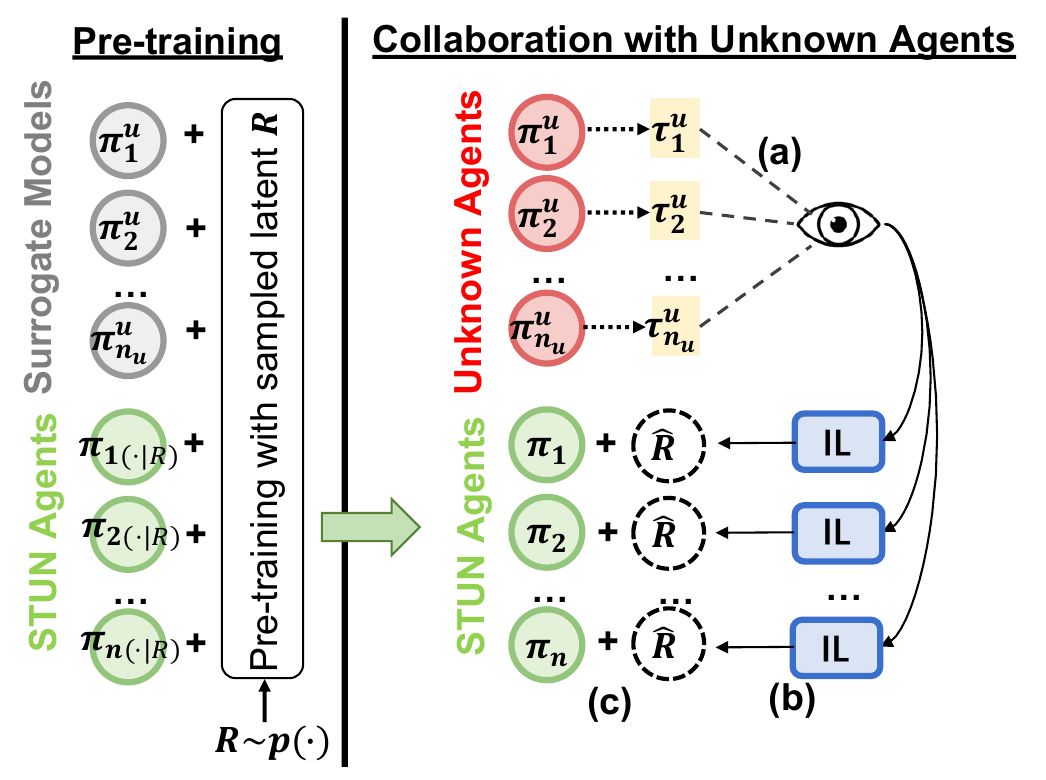}}
\vspace{-0.2in}
\caption{An illustration of our proposed framework. STUN agents $\pi_i(\cdot|R)$ are pre-trained using surrogate agent models with sampled latent rewards $R$. To collaborate with unknown agents, they use inverse learning (step b) on the observed trajectories $\{\tau^u_i\}$ of the unknown agents (step a) and perform a zero-shot policy adaptation based on an unbiased estimate $\hat{R}$ (step c).}
\vspace{-0.4in}
\label{fig:pipeline}
\end{center}
\end{wrapfigure}

As illustrated in Figure~\ref{fig:pipeline}, to support synergistic teaming, a team of STUN agents is deployed in a shared task environment to collaborate with unknown agents. The STUN agents can observe the trajectories $\{\tau^{u}\} =\{(o^u_i,a^u_i)\}_{i=1}^n$ of the unknown agents (defined as a sequence of their joint observations
$o^u_i\in \mathcal{O}^u=O_1\times O_2\times\cdots\times O_{n_u}$ and joint actions $a^u_i\in \mathcal{A}^u=A_1\times A_2\times\cdots\times A_{n_u}$). However, they do not have access to the latent reward $R$ that is controlled only by the unknown agents. To avoid the overhead of re-training or transfer-learning, we propose a zero-shot policy adaptation framework. The key idea is to pre-train a class of goal-conditioned policies $\pi(a|o,R)$ for the STUN agents, which are conditioned on potential reward functions $R$. This pre-training is supported by the use of surrogate unknown agent models with respect to randomly sampled latent reward function $R$. 

The pre-trained STUN agents (with goal-conditioned policies $\{\pi_i(a_i|o_i,R)\}$) are then deployed in a shared task environment to collaborate with unknown agents and to support common goals/objectives. We propose a kernel-density Bayesian inverse learning (KD-BIL) algorithm to obtain the posterior distribution of latent reward function $\mathbb{P}(R|\{(o^u_i,a^u_i)\}_{i=1}^n)$ from the observed trajectories of $\{\tau^{u}\} =\{(o^u_i,a^u_i)\}_{i=1}^n$. The proposed inverse learning algorithm uses the kernel-density method to estimate the posterior and thus eliminates the need to refit policies for each sampled reward function. It allows obtaining the posterior estimates from even limited observation (i.e., with small $n$) or when the reward is time-varying (i.e., updating the latest $n$ observations).

We show that a maximum a posteriori (MAP) of latent reward cannot ensure convergence of the Bellman equation to achieve optimal expected return $G = \sum^{H}_{t}\gamma^{t}R^{t}$. Instead, an unbiased estimate $\tilde{R}$ is proven to be necessary for learning the optimal collaborative policy. Based on this result, we leverage the pre-trained, goal-conditioned policies $\{\pi_i\}$ and perform a zero-shot policy adaptation using the unbiased estimate, i.e, $\{\pi_i(a_i|o_i,\tilde{R})\}$. The proposed policy adaptation requires no re-training or re-learning (thus zero-shot), while ensuring optimality of the adapted policy.

\subsection{Active Goal Inference with Inverse Learning}

Our proposed KD-BIL is a method for approximating the reward probability distribution using kernel density estimates. It not only allows an efficient estimate of the unknown agents' reward (which is often associated with substantial uncertainty) but also supports sample-efficient computations using limited observation data. Specifically, 
we set up a training dataset $\mathbb{D}$ by sampling $m$ demonstrations, either from available trajectories of known agents with observable reward or by training surrogate unknown agents using sampled reward functions.  
This dataset consists of $m$ 3-tuple demonstrations $\{ (o_{j},a_{j},R_{j}) \}_{j=1}^{m}$, where $R_{j}$ is the reward function used to generate the demonstrations $(o_{j},a_{j})$. Using either known agents or surrogate agent models, we construct the training dataset that contains demonstrations of various potential behaviors.

Given observed unknown agent trajectories $(o_{i}^{u},a_{i}^{u})$ of size $n$, we can now estimate the posterior $p_{m}(R|o_{i}^{u},a_{i}^{u})$ by formulating conditional density $\hat{p}_{m}(o_{i}^{u},a_{i}^{u}|R)$ with respect to the training dataset and our choice of kernel density function $K(\cdot,\cdot)$. 

Using demonstrations in the training dataset, the conditional density for a state-action pair $(o_i^u,a_i^u)$ given a latent reward function $R$ is
\begin{equation}
\hat{p}_{m}(o_{i}^{u},a_{i}^{u}|R)\propto \sum\limits_{j=1}^{m} \frac{K((o_{i}^{u},a_{i}^{u}),(o_{j},a_{j}))\cdot K_{R}(R,R_{j})}{\sum_{l=1}^{m}K_{R}(R,R_{l})}
\end{equation}

where $K(\cdot,\cdot)$ and $K_R(\cdot,\cdot)$ are two different kernel density functions for the state-action pair and for the reward, respectively. The proposed KD-BIL method works with any kernel functions, such as the Gaussian kernel and the Matern kernel~\cite{KD_BIRL}. We consider the Gaussian kernel function in the following derivations. This implies that the conditional density $\hat{p}_{m}(o_{i}^{u},a_{i}^{u}|R)$ is given by:

\begin{equation}
\begin{aligned}
\hat{p}_{m}(o_{i}^{u},a_{i}^{u}|R)
\propto \sum\limits_{j=1}^{m} \frac{e^{-d_{s}((o_{i}^{u},a_{i}^{u}),(o_{j},a_{j}))^{2}/(2h)}e^{-d_{r}(R,R_{j})^{2}/(2h^{\prime})}}{\sum_{l=1}^{m}e^{-d_{r}(R,R_{l})^{2}/(2h^{\prime})}}
\end{aligned}
\end{equation}

where $d_{s}:(\mathcal{O}\times \mathcal{A})\times (\mathcal{O} \times \mathcal{A}) \rightarrow \mathbb{R}$ is a distance metric to compare $(o,a)$ tuples, $d_{r}:R\times R \rightarrow \mathbb{R}$ is a distance metric to compare reward functions, and $h,h^{'}>0$ are smoothing hyperparameters. We note that these hyperparameters can be further optimized using the training dataset, e.g., similar to optimizing surrogate models in Bayesian Optimization~\cite{ramachandran2007bayesian}. Next, we obtain the following posterior estimate of latent reward.

\begin{lemma}
\label{lem:KD_BIRL}
The estimated posterior of the unknown agent reward is given by
\begin{equation}
\begin{aligned}
 \hat{p}_{m}(R|\{o_{i}^{u},a_{i}^{u}\}_{i=1}^{n})  
\propto  p(R) \prod\limits_{i=1}^{n} \sum\limits_{j=1}^{m} \frac{e^{-d_{s}((o_{i}^{u},a_{i}^{u}),(o_{j},a_{j}))^{2}/(2h)}e^{-d_{r}(R,R_{j})^{2}/(2h^{\prime})}}{\sum_{l=1}^{m}e^{-d_{r}(R,R_{l})^{2}/(2h^{\prime})}}
\label{eq:posterior}
\end{aligned}
\end{equation}
\end{lemma}

Importantly, we note that this conditional density $\hat{p}_{m}$ can be easily computed from the training dataset and using kernel density functions. There is no need to refit policies or perform value iterations to evaluate the posterior for a given reward function $R$. This drastically reduces the computational complexity compared to existing Bayesian IRL algorithms~\cite{KD_BIRL,ramachandran2007bayesian,choi2012nonparametric,chan2021scalable} and thus makes it possible to perform Bayesian inverse learning for the unknown agent's latent reward, even in complex environments with large state spaces (as shown in our evaluation on SMAC~\cite{SMAC}). Further, $p(R)$ in Equation~(\ref{eq:posterior}) denotes the prior distribution of reward functions. It can be a uniform distribution or estimated from available unknown agent statistics. As $m \rightarrow \infty$, it is shown that $\hat{p}_{m}$ converges to the true likelihood of reinforcement learning policies and the posterior converges to the true posterior~\cite{van2000asymptotic}.

In practice, we can consider a general representation of the latent reward function, i.e., $R_{\mathcal{B}}(s,a)$ with latent parameters $\mathcal{B} \in \mathbb{R}^k$, which $k$ is the latent dimension of $\mathcal{B}$. This representation captures latent reward that can be expressed as a linear function $R_{\mathcal{B}}(s, a) = \mathcal{B}^{T}\mathbf{R}(s, a)$ of possible underlying components $\mathbf{R}=(R_1, R_2,\ldots, R_k)$ of the unknown agents' potential objectives, as well as more general forms of reward functions that are defined through a neural network $R_{\mathcal{B}}(s,a)$ parameterized by $\mathcal{B}$. For instance, in our evaluations using the MPE environment~\cite{MPE}, we can construct a mixing network (e,g., a single-layer neural network parameterized by $\mathcal{B}$) to combine underlying components such as greedy, safety, cost, and preference, into a more complex and realistic reward function representation for goal inference.  Thus, active goal inference in this paper aims to estimate the latent reward parameters $\mathcal{B}$ from observed unknown agents' trajectories $\{(o^u_i,a^u_i)\}_{i=1}^n$, using the proposed KD-BIL method, by estimating posterior $\hat{p}_{m}(R_{\mathcal{B}}|\{o_{i}^{u},a_{i}^{u}\}_{i=1}^{n})$ over the latent reward parameters $\mathcal{B}$ instead. This approach enables efficient estimate of complex reward functions in MPE and SMAC environments.

\subsection{Zero-shot Policy Adaptation}

With the posterior estimates of latent reward, it would be tempting to consider the MAP estimate $\hat{\mathcal{B}}^* = \arg \max_{\hat{\mathcal{B}}} \hat{p}_{m}(R_{\hat{\mathcal{B}}}|\{o_{i}^{u},a_{i}^{u}\}_{i=1}^{n})$ and use it directly to adapt collaborative AI agent policies. However, as shown in our next theorem, unbiased estimates of the latent reward $R_{\tilde{\mathcal{B}}}$ are needed to ensure the convergence of Bellman equations to the optimal values. Furthermore, directly employing the estimated reward for re-training and re-learning the collaborative AI agent policies on the fly can result in significant overhead and unstable teaming performances. To this end, we propose a novel zero-shot policy adaptation for teaming with unknown agents. It pre-trains a set of goal-conditioned policies $\{\pi_i(a_i|o_i,{R})\}$ for the collaborative AI agents (by leveraging surrogate unknown agent models) and then makes a zero-shot policy adaptation using unbiased reward estimates.

To establish optimality of the proposed approach, we consider a Q-learning algorithm (e.g., {~\cite{watkins1992q} that are often employed for theoretical analysis) over the joint action and state space and under the unbiased reward estimates $R_{\tilde{\mathcal{B}}}$. Thus, the Q-values, $Q^{\pi}_{\mathcal{B}}(s,a) = \mathbb{E}_{\pi}[\sum_{t}^H \gamma^t R^t |S_{t}=s,A_{t}=a, R^t\sim R_{\mathcal{B}}]$, are now defined with respect to the reward estimates $R_{\tilde{\mathcal{B}}}$. We show that unbiased reward estimates are sufficient to ensure convergence to the optimal Q-values that are achieved by having the actual reward. 
\begin{theorem}
\label{thm:q_learning}
(Unbiased Estimates Ensures Optimality.) Given a finite MDP, denoting as $\mathcal{M}$, the Q-learning algorithm with unbiased estimate rewards satisfying $\mathbb{E}[R_{\tilde{\mathcal{B}}}]=R_{\mathcal{B}}$, given by the update rule,
\begin{equation}
\label{eql:q_learning}
\begin{aligned}
    Q_{\tilde{\mathcal{B}},t+1}(s_{t},a_{t}) 
    = (1-\alpha_{t})Q_{\tilde{\mathcal{B}},t}(s_{t},a_{t})  +\alpha_{t}[R_{\tilde{\mathcal{B}}}+\gamma \max\limits_{b\in \mathcal{A}}Q_{\mathcal{B},t}(s_{t+1},b)]
\end{aligned}
\end{equation}
converges w.p.1 to the optimal Q-function as long as $\sum_{t}\alpha_{t} = \infty$ and $\sum_{t}\alpha_{t}^{2}<\infty$

\end{theorem}

Note that the right hand of Equation~(\ref{eql:q_learning}) relies on unbiased estimate reward $R_{\tilde{\mathcal{B}}}(s,a)$. Theorem~\ref{thm:q_learning} states that collaborative agent policies will converge to optimal $w.p.1$ when replacing the actual rewards with unbiased estimates. 

\textbf{Estimating unbiased rewards.}
%
%
Next, we obtain an unbiased reward estimate $R_{\tilde{\mathcal{B}}}$. For linear reward function $R$ over the latent parameters, the problem is equivalent to obtaining an unbiased estimate of the latent parameters satisfying $\mathbb{E}[\tilde{\mathcal{B}}] = \mathcal{B}$. Let $\hat{B}$ be the posterior distribution of the latent parameters. We can estimate unbiased $\tilde{\mathcal{B}}$ from $\hat{B}$.
\begin{lemma}
\label{lem:linear_B}
If the reward function is linear over the latent parameters, $\tilde{\mathcal{B}} = \mathbb{E}[\mathcal{B}|\hat{\mathcal{B}}]$ gives
an unbiased estimate, i.e., $\mathbb{E}[R_{\mathcal{\tilde{B}}}(s,a)] = R_{\mathcal{B}}(s,a)$.
\end{lemma}
Since the conditional distribution $d_{\mathcal{B}|\hat{\mathcal{B}}}$ may not be available, we can leverage a neural network during the pre-training stage to estimate it -- as actual $\mathcal{B}$ of the surrogate models and posterior estimates $\hat{\mathcal{B}}$ both are available during pre-training. Another idea is to use the posterior distribution as an approximation, i.e.,   
$d_{\mathcal{B}|\hat{\mathcal{B}}} \sim \hat{p}_{m}(\cdot|\{o_{i}^{u},a_{i}^{u}\}_{i=1}^{n})$. Our evaluations show that this method allows efficient approximation of the unbiased estimate with negligible error. 
For general non-linear reward functions, we denote $R^{-1}$ as an inverse in the sense that $R^{-1}(R_{\mathcal{B}})$ recovers the underlying parameter ${\mathcal{B}}$ of the reward function $R_{\mathcal{B}}$. An unbiased estimate can then be obtained as follows.
\begin{lemma}
\label{lem:noliner_B}
$\tilde{\mathcal{B}} = R^{-1}(\mathbb{E}[R_{\mathcal{B}}|\hat{\mathcal{B}}] )$ is an unbiased estimate of $\mathcal{B}$, i.e., $\mathbb{E}[R_{\tilde{\mathcal{B}}}]=\mathbb{E}[R_{\mathcal{B}}]$. 
\end{lemma}
In practice, we can always train a neural network to recover $\tilde{\mathcal{B}}$ from the posterior distribution of $\hat{\mathcal{B}}$, with the goal of minimizing the resulting bias of the recovered reward function $R_{\tilde{\mathcal{B}}}$. Let $g_\zeta$ with parameter $\zeta$ be the neural network. We consider the estimate $\tilde{\mathcal{B}}=g_\zeta(\hat{\mathcal{B}})$ to minimize a loss function the mean square error of the resulting reward bias: 
\begin{eqnarray}
L_\zeta=\mathbb{E}[(R_{\tilde{\mathcal{B}}}-R_\mathcal{B})^2], \ {\rm s.t.} \ \tilde{\mathcal{B}} = g_\zeta(\hat{\mathcal{B}})
\end{eqnarray}
Such $g_\zeta$ can be optimized during the pre-training stage (as shown in Figure~\ref{fig:pipeline}) using the actual $R_\mathcal{B}$ of the surrogate models and the posterior $\hat{\mathcal{B}}$ from proposed KD-BIL.

\noindent {\bf Zero-shot adaptation with pre-training.} 
We pre-train a set of goal-conditioned policies $\{\pi_i(a_i|o_i,{\mathcal{B}})\}$ -- which are now conditioned on the latent reward function parameters ${\mathcal{B}}$ instead -- for the collaborative AI agents. The pre-training is illustrated in Figure~\ref{fig:pipeline}, where surrogate models of the unknown agent are leveraged. We employ MARL to train the surrogate models and the collaborative AI agent policies, using reward signals $R_{\mathcal{B}}$ with randomly sampled parameters $\mathcal{B}\sim p(\cdot)$. Thus, a zero-shot policy adaptation can be achieved during the teaming stage by feeding the unbiased estimate $\tilde{\mathcal{B}}$ (of unknown agent reward) into the goal-conditioned policies, i.e., $\{\pi_i(a_i|o_i,\tilde{\mathcal{B}})\}$. It ensures optimal teaming performance with unknown agents.

The pseudo-code of our proposed STUN framework can be found in Appendix~A. In particular, the pre-training of goal-conditioned policies (together with surrogate models) can leverage any MARL algorithms to maximize the expected return $J_{\mathcal{B}}(\theta)$ under-sampled reward:
\begin{equation}
    J_{\pi,\mathcal{B}}(\theta) = \mathbb{E}_{s_{0}\sim\mu, s, a} [V^{\pi_{\theta}}_{\mathcal{B}}(s_{0})]
\end{equation}
where $s_{0}$ is drawn from the initial state distribution. In this paper, we use policy gradient algorithms to pre-train the goal-condition policies. 

\begin{lemma}
\label{lem:latentpolicy}
(Latent Style Policy Gradient for pre-training). 
\begin{equation}
     \nabla_{\theta} J_{\pi,\mathcal{B}}(\theta) = \mathbb{E}_{\pi_{\theta}}[Q^{\pi_{\theta}}_{\mathcal{B}}(s,a)\nabla_{\theta}\log \pi_{\theta}(a|o,{B})]
\end{equation}
\end{lemma}

\begin{figure*}[h]
\vskip 0.2in
\centering
  \subfigure[Interpreting teaming behavior.]{\includegraphics[width=0.32\textwidth]{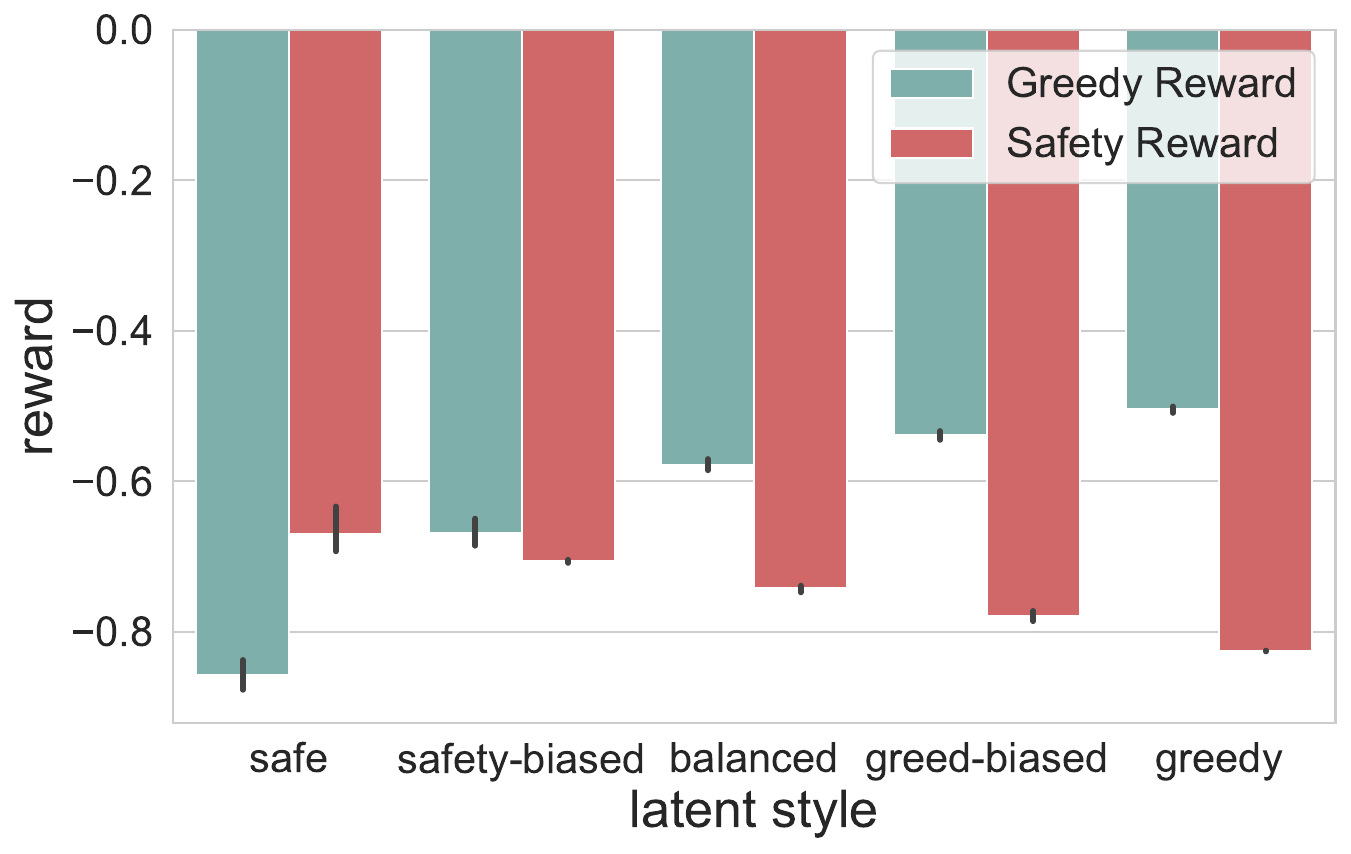}
  \label{fig:mpe_interpretability}
  }\hfill
  \subfigure[Teaming with changing unknown agents.]{\includegraphics[width=0.32\textwidth]{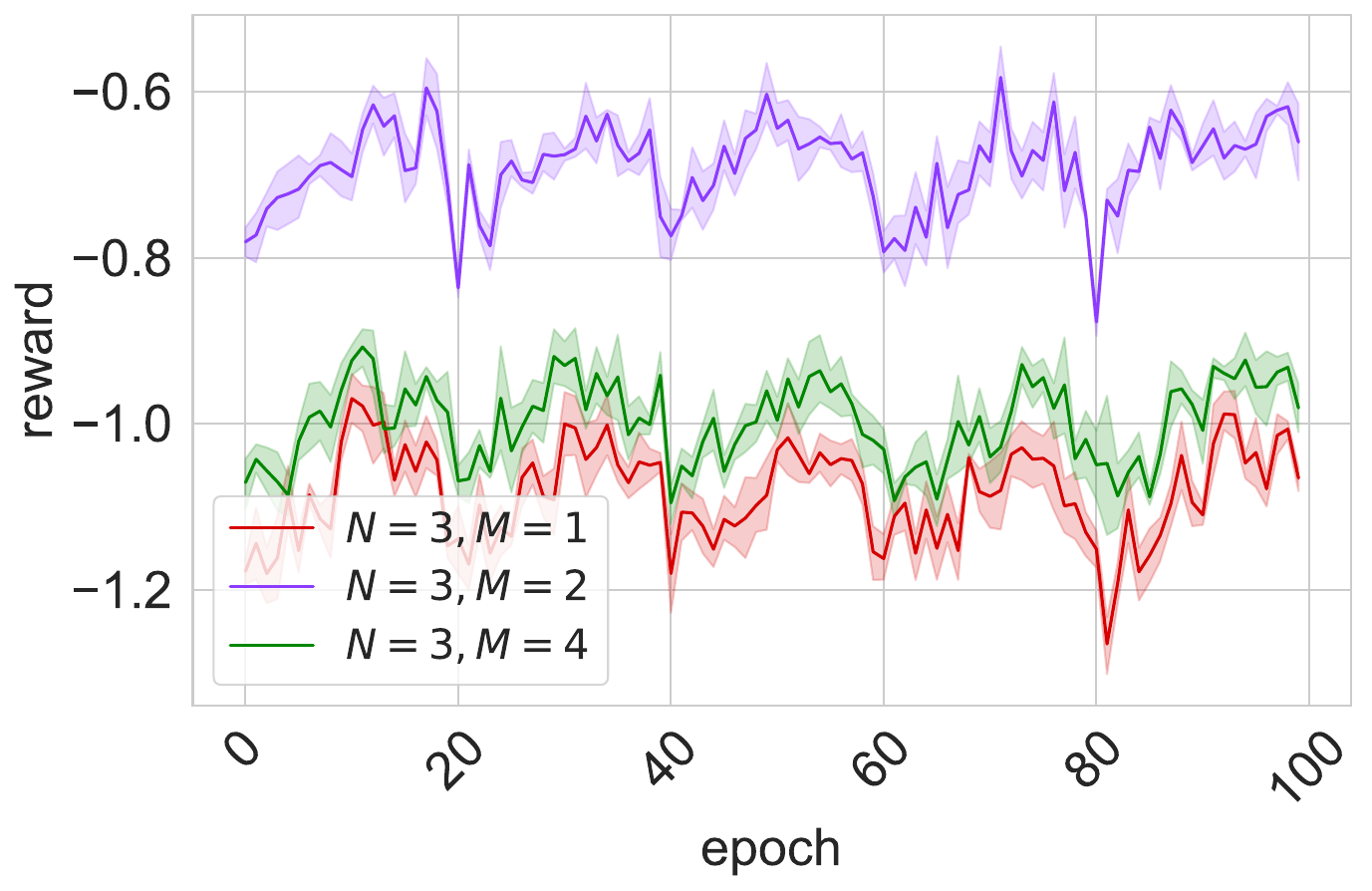}
  \label{fig:mpe_dynamic}
  }\hfill
  \subfigure[Ablation and scalability study.]{\includegraphics[width=0.32\textwidth]{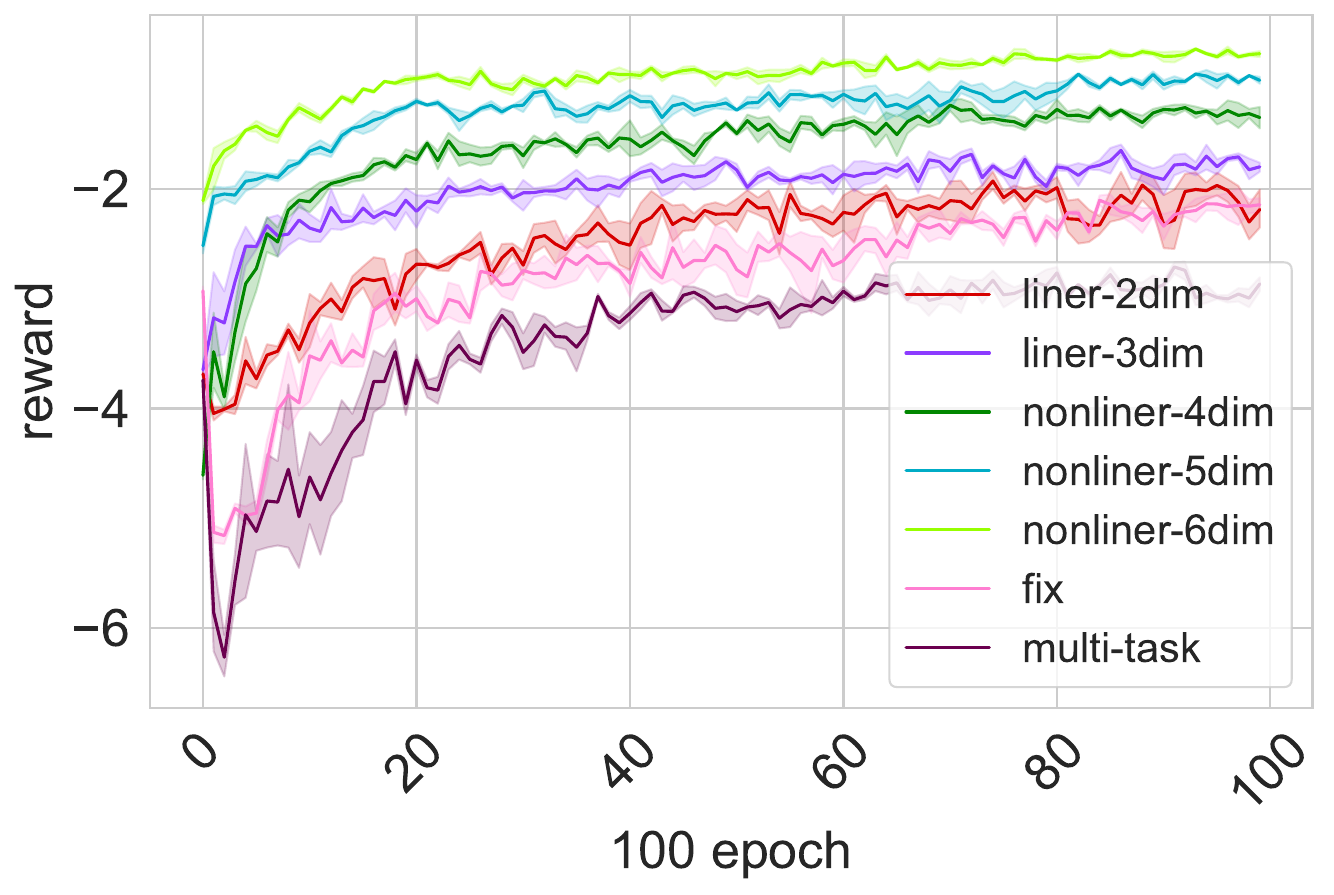}
  \label{fig:mpe_ablation}
  }\hfill\vspace{-1.5ex}
  \caption{(a) Illustration of the underlying reward tradeoff when STUN agents team up with unknown agents ranging from playing safe to being greedy. (b) Ability of STUN agents to quickly reasoning/infering the time-varying reward of the unknown agents (changing every 20 epochs) and then performing zero-shot policy adaptation on the fly. (c) Ablation studies showing the impact of different design modules, as well as robust performance of STUN under unknown reward function with increasing complexity (e.g., increasing from 2 to 6 dimensions of reward components and using nonlinear mixing functions).}
  \label{fig:mpe}
\vskip -0.1in
\end{figure*}

\begin{figure*}[h]
\centering
  
  \subfigure[3s\_vs\_5z]{\includegraphics[width=0.32\textwidth]{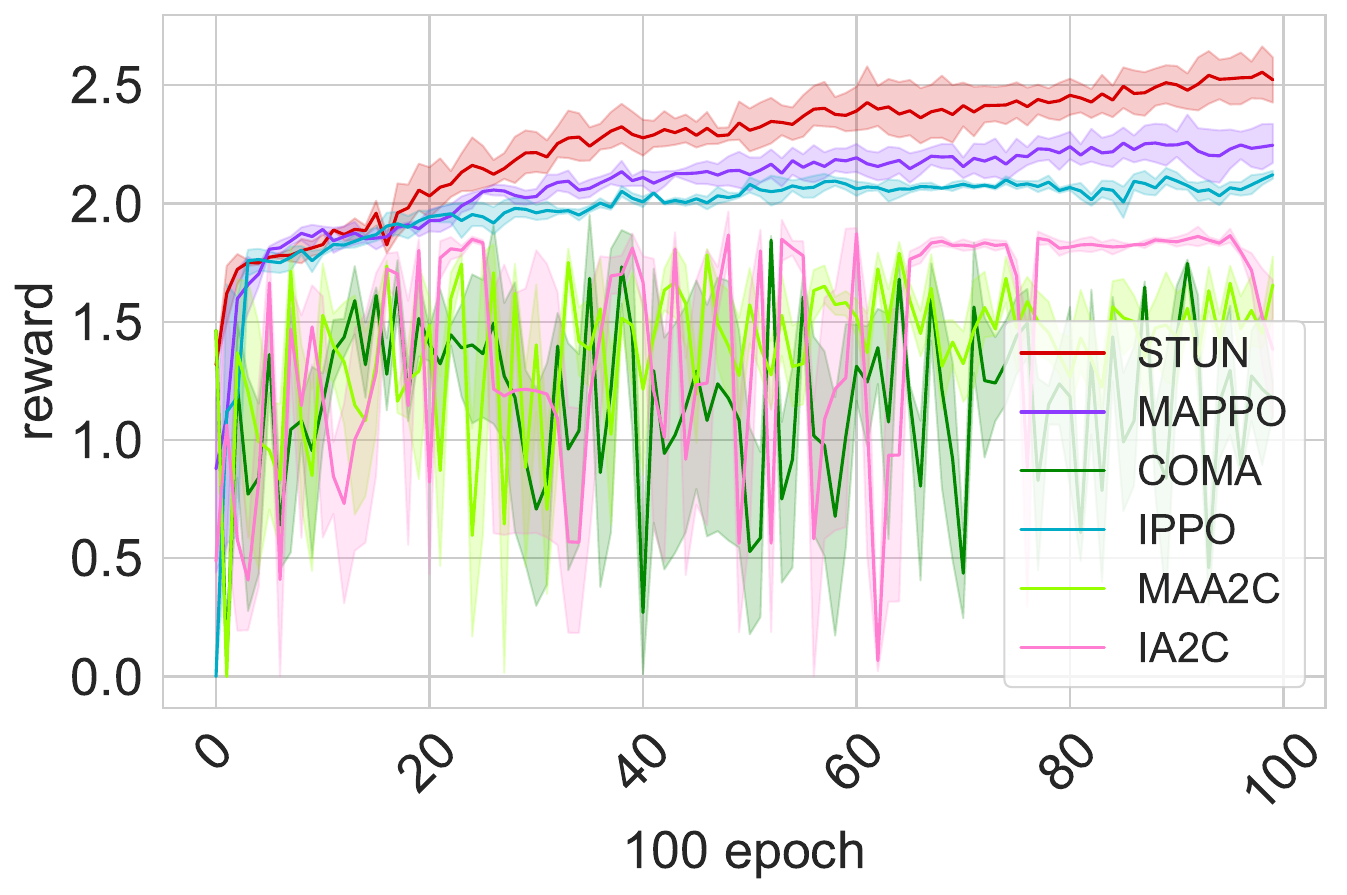}
  \label{fig:smac_3s_vs_5z}
  }\hfill
  \subfigure[5m\_vs\_6m]{\includegraphics[width=0.32\textwidth]{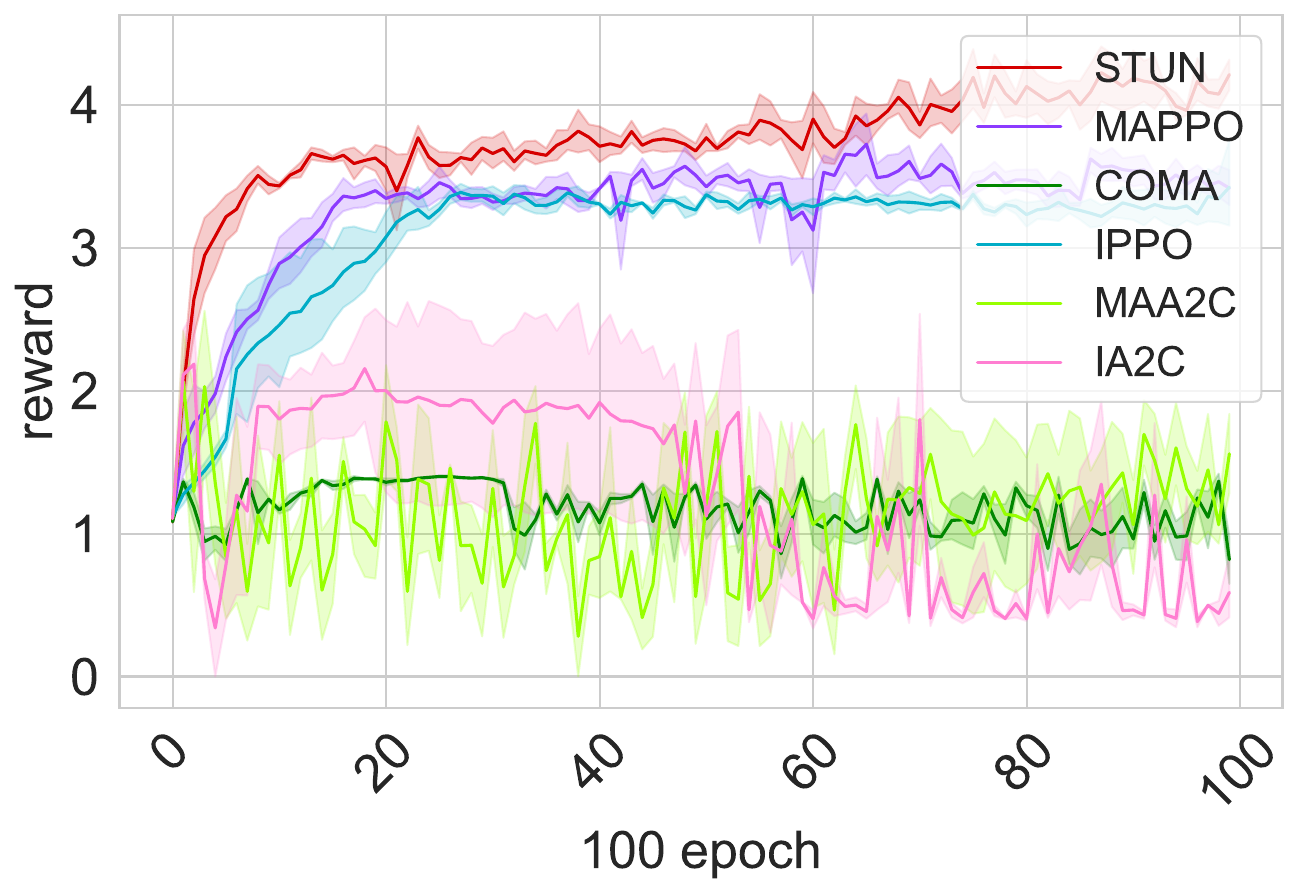}
  \label{fig:smac_5m_vs_6m}
  }\hfill
  \subfigure[6h\_vs\_8z]{\includegraphics[width=0.32\textwidth]{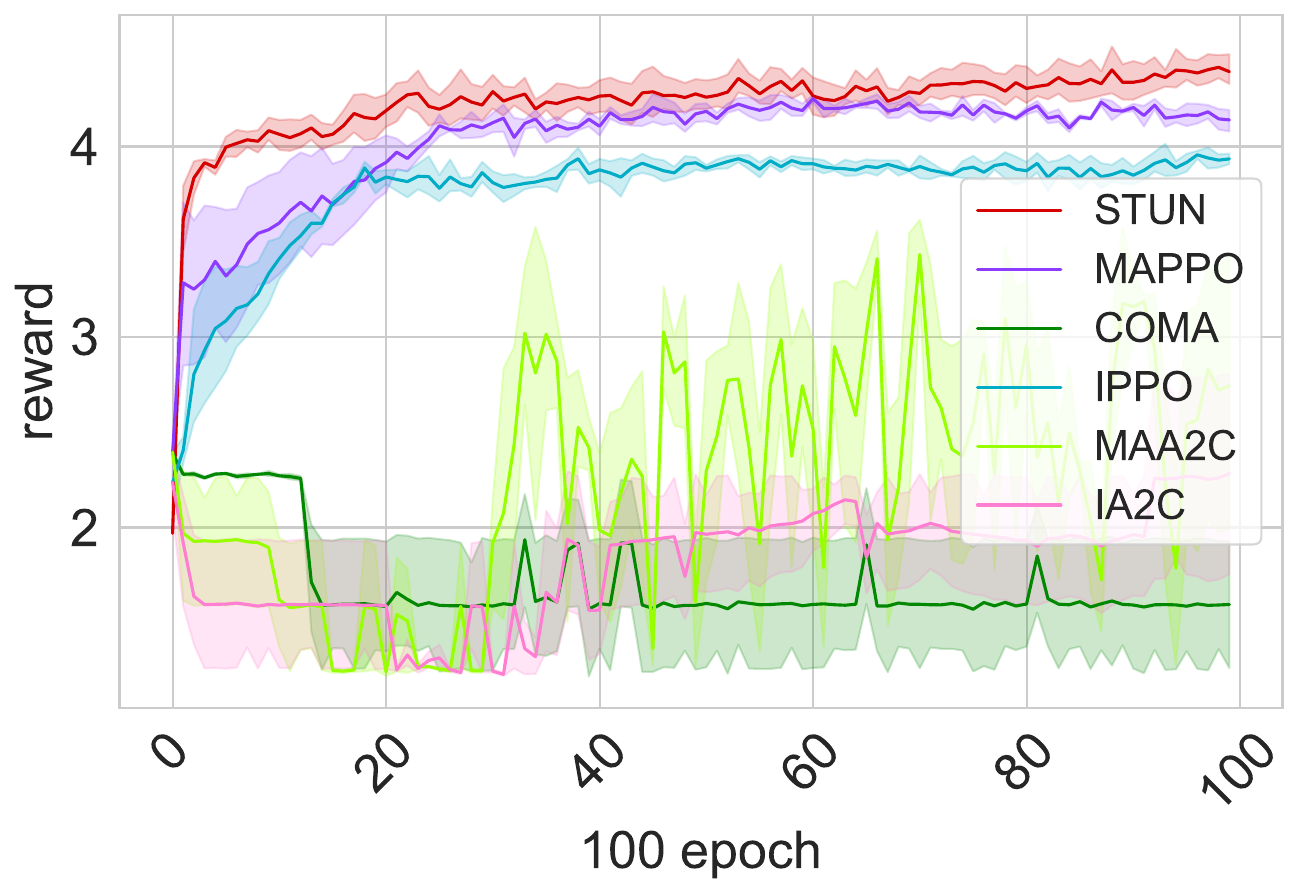}
  \label{fig:smac_6h_vs_8z}
  }\hfill
  \\
  \subfigure[27m\_vs\_30m]{\includegraphics[width=0.32\textwidth]{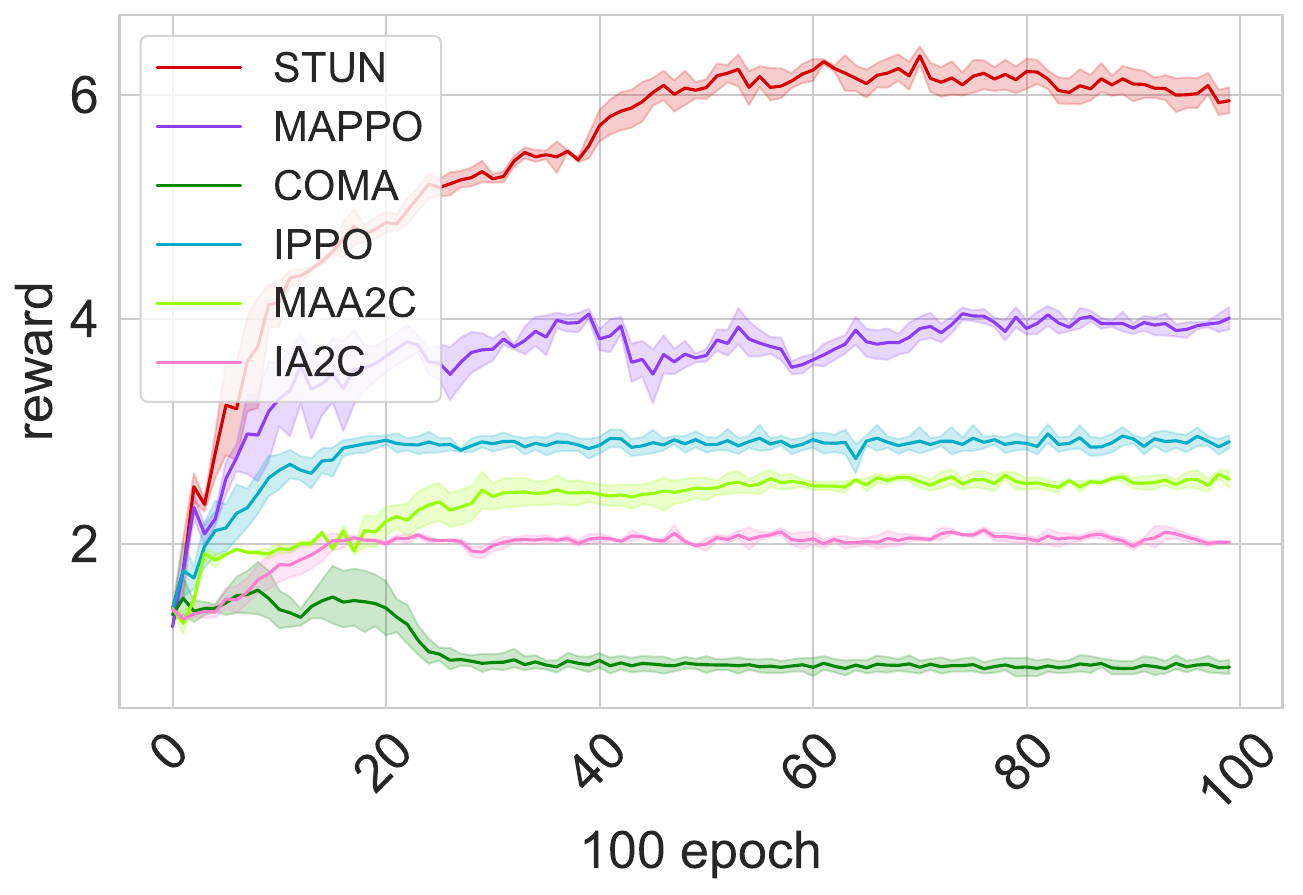}
  \label{fig:smac_27m_vs_30m}
  }\hfill
  \subfigure[corridor]{\includegraphics[width=0.32\textwidth]{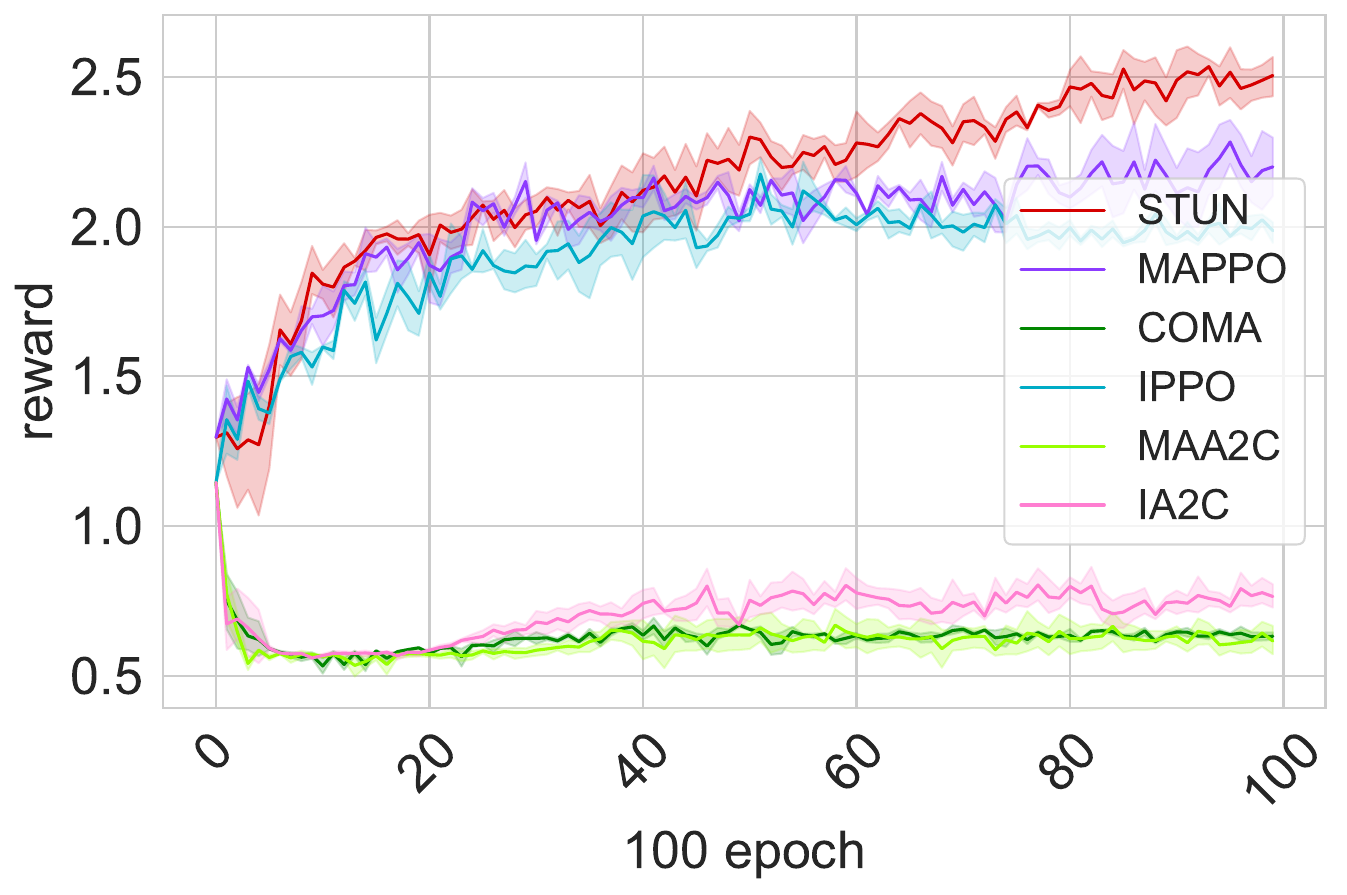}
  \label{fig:smac_corridor}
  }\hfill
  \subfigure[MMM2 (hard)]{\includegraphics[width=0.32\textwidth]{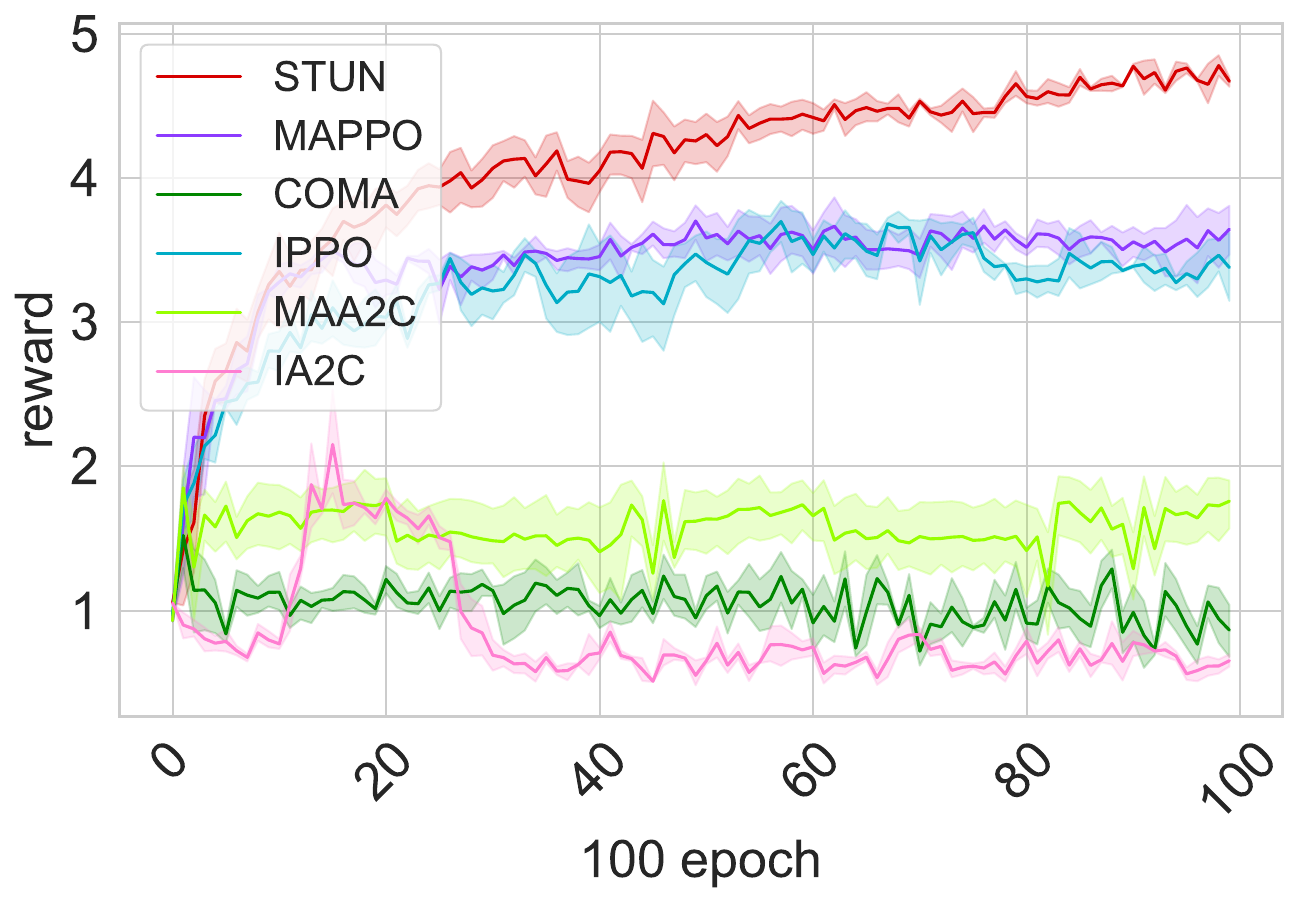}
  \label{fig:smac_MMM2}
  }\hfill\vspace{-1ex}
   \caption{Performance comparison of our proposed STUN agents and selected baselines on redesigned SMAC tasks.}
  \label{fig:smac}
\end{figure*}

\section{Experiments}

We redesigned two multi-agent simulation environments based on multiagent-particle-envs(MPE)~\cite{MPE} and SMAC~\cite{SMAC} to create collaborative teaming tasks with unknown agents.
These environments include blue (friendly) and red (adversarial) teams. The red team is controlled by built-in policies within the environments, while the blue team consists of both unknown agents and collaborative AI agents. We consider different methods for creating the unknown agents, such as training with randomly-generated latent reward parameters and using available agents from popular MARL algorithms like MAPPO, IPPO, COMA, and IA2C ~\cite{MAPPO,IPPO,COMA,IA2C}. We deploy STUN agents (and other baseline agents obtained by multi-task learning) alongside these unknown agents and evaluate the teaming performance in each collaborative task. Related code and implementation details can be found at GitHub \textbf{(see supplementary Files)}.

\begin{table*}[h!]
\label{tab:smac}
\centering
\resizebox{0.985\textwidth}{!}{%
\begin{tabular}{@{}cclccclccccc@{}}
\toprule\toprule
\multirow{2}{*}{Unknown Agents} & Synergistic Agents &  & \multicolumn{3}{c}{Fixed-Behavior Agents} &  & \multicolumn{5}{c}{Multi-task Agents} \\ \cmidrule(l){2-12} 
           & \textbf{STUN} &  & FBA-C & FBA-B & FBA-A         &  & MAPPO & IPPO & COMA          & MAA2C & IA2C \\ \midrule
FBA-C      & \textbf{1.10} &  & 1.04  & 1.04  & 0.99          &  & 0.88  & 0.85 & 0.84          & 0.69  & 0.68 \\
FBA-B      & \textbf{2.38} &  & 2.06  & 2.36  & 2.19          &  & 1.99  & 1.84 & 1.83          & 1.71  & 1.73 \\
FBA-A      & 3.73          &  & 2.88  & 3.55  & \textbf{3.88} &  & 3.08  & 2.94 & 2.68          & 2.35  & 2.67 \\ \midrule
MAPPO      & \textbf{2.43} &  & 2.04  & 2.05  & 2.08          &  & 2.11  & 1.97 & 1.80          & 1.87  & 1.56 \\
IPPO       & \textbf{2.25} &  & 1.82  & 2.22  & 2.19          &  & 1.95  & 2.22 & 2.06          & 1.84  & 1.87 \\
COMA       & 2.12          &  & 1.62  & 1.86  & 1.95          &  & 1.98  & 1.81 & \textbf{2.18} & 1.55  & 1.58 \\
MAA2C      & \textbf{2.22} &  & 1.58  & 1.97  & 2.05          &  & 1.25  & 1.83 & 1.79          & 2.13  & 2.12 \\
IA2C       & \textbf{2.08} &  & 1.73  & 1.89  & 1.87          &  & 1.84  & 1.62 & 1.55          & 1.43  & 1.97 \\ \midrule
Average      & \textbf{2.29} &  & 1.85  & 2.12  & 2.15          &  & 1.89  & 1.89 & 1.84          & 1.70  & 1.77 \\
Normalized & \textbf{99.2} &  & 81.1  & 91.6  & 92.1          &  & 81.5  & 81.7 & 80.2          & 73.7 & 76.7 \\ \bottomrule\bottomrule
\end{tabular}%
}
\caption{Experimental results of average reward on \texttt{3s\_vs\_5z} map of the redesigned SMAC environment. STUN and selected baselines (after training) are each teamed up with 8 different unknown agents, respectively. 
Each row represents the teaming performance with a different unknown agent. STUN achieves nearly optimal performance in nearly all scenarios, demonstrating its robust performance.}

\label{tab:my-table}
\end{table*}

\subsection{Multi-Agent Particle Environment}

We first consider Predator-Prey from the MPE environment, which is a partially observable multi-agent environment that involves $N$ AI-controlled blue agents and $M$ adversary agents. Half of the blue agents are unknown agents with latent rewards, while the rest of the blue agents are collaborative AI agents. 
Meanwhile, the adversaries follow a fixed strategy: they move towards the nearest agent, and different adversaries will choose different targets. 

To create unknown agents with diverse behaviors/objectives, we consider four methods of creating reward components and combine them into more complex reward functions $R_{\mathcal{B}}(s, a)$ using either a linear function or a non-linear mixing network (e.g., a single-layer neural network) with latent parameters $\mathcal{B}$. We consider  \textbf{ (i) Greedy Reward:} As Preys get closer to each other, they are less greedy in terms of exploration, thus resulting in a negative reward. \textbf{(ii) Safety Reward:} Prey attempts to evade Predators and receives a negative reward proportional to the distance.  \textbf{(iii) Cost Reward:} Movement by the Prey consumes energy, so when the Prey moves, it also receives a negative reward.  \textbf{(iv) Preference Reward:} Different weights can be assigned to Predator-Pray pairs in the other classes to reflect individual preferences/importance. We note that the combination of these methods allows the creation of complex reward functions with many dimensions. 

We evaluate the performance of STUN agents teaming up with unknown agents and focus on two key aspects: {(1) Adaptability:} evaluating whether trained STUN agents can maintain high teaming performance when collaborating with new, unknown agents or agents with time-varying behaviors/objectives. (2) Interpretability of behavior: Assessing how collaborative agents' behaviors vary under different unknown agents with latent $\mathcal{B}$. 
We also perform ablation studies to verify the impact of (i) zero-shot adaptation using goal-conditioned policies and (ii) active goal inference of our proposed design, as well as the impact of high-dimensional, linear, and nonlinear reward functions. 

\textbf{Teaming behavior interpretation.} To better interpret STUN agent behaviors, we focus on Greedy Rewards and Safety Rewards in this experiment. While more details are provided in~Appendix C.1.2, we show in Fig.~\ref{fig:mpe_interpretability}(a) the achieved tradeoff between the two reward components when collaborating with unknown agents of different latent $\mathcal{B}$. As the unknown agents tend to move from playing safe (i.e., staying away from predators) to being greedy (i.e., more aggressively exploring), STUN agents adapt their policies and also become more greedy -- as shown by diminishing safety return and increasing greedy return.
More teaming analysis and illustrations are provided in Appendix C.1.2.

\textbf{Collaborating with changing unknown agents.}
During the teaming/execution stage, we deploy trained STUN agents alongside unknown agents with changing behavior. Specifically, the unknown agents vary their policies at the beginning of every 20 epochs. This requires STUN agents to continually reason/infer the time-varying reward of the unknown agents and then perform zero-shot policy adaptation on the fly. Fig.~\ref{fig:mpe_interpretability}(b) shows that STUN agents can swiftly adapt their policies in just 5-10 epochs (with goal inference and zero-shot policy adaptation) and ramp up teaming performance in different environments with $M$ adversaries.

\textbf{Ablation studies.} We now perform an ablation study to (i) remove the zero-shot policy adaptation in STUN agents by instead performing additional online reinforcement learning using the inferred reward and (ii) remove the active goal inference by conditioning STUN agents on fixed reward parameters -- labeled ``multi-task" and ``fix" respectively in Fig.~\ref{fig:mpe_interpretability}(c). Significant performance degradations are observed compared to STUN agents labeled ``nonlinear-4dm". For scalability, in Fig.~\ref{fig:mpe_interpretability}(c), we further vary the dimensions of underlying reward components from 2 to 6 and evaluate STUN agents over both linear and non-linear reward functions (e.g., soft-max and single-layer network with parameters $\mathcal{B}$). The numerical results demonstrate STUN agents' robust teaming performance with increasingly complex unknown reward structures.

\subsection{StarCraft Multi-Agent Challenge}

In this section, we perform extensive evaluations of the proposed framework on SMAC tasks (e.g., hard and super-hard maps) and compare it with a range of baseline algorithms. Note that to create unknown agents with different latent rewards, we have redesigned the SMAC environment\footnote{Standard SMAC environment considers only winning rate as the reward, which is insufficient for creating diverse unknown agents with latent rewards for our evaluation.} to consider two broad classes of rewards: \textbf{Conservative Rewards} that 
are represented by the health values of surviving friendly blue-team agents and \textbf{Aggressive Rewards} that are represented by the total damage inflicted on adversarial red-team agents. This design allows us to create diverse unknown agents with different latent reward functions and play styles, ranging from conservative to aggressive as parameterized by the latent $\mathcal{B}$. Teaming performance is measured using the achieved (latent) reward. All other environment settings remain the same as standard SMAC. Detailed information on our settings and training configurations, like hyperparameters used, can be found in the Appendix.

We consider 6 maps selected from SMAC with varying levels of difficulty and create 7 types of unknown agents either by training with fixed unknown behaviors or by directly using agents from popular MARL algorithms, including MAPPO, IPPO, COMA, and IA2C ~\cite{MAPPO,IPPO,COMA,IA2C}. We deploy trained STUN agents in these collaborative tasks against each type of unknown agents and compare the performance to a number of baselines such as optimal fixed-behavior agents (e.g., with conservative (FBA-C), balanced (FBA-B), and aggressive (FBA-A) play styles) and collaborative agents that employ multi-task learning by randomly sampling the unknown agents' latent parameters. In the following evaluations, we will demonstrate: (1) The proposed KD-BIL can accurately infer latent reward parameters $\mathcal{B}$; (2) STUN agents can efficiently team up with unknown agents and outperform baselines on various SMAC tasks; and (3) STUN agents demonstrate robust performance with diverse unknown agent behaviors/objectives.

\textbf{Evaluating goal inference against ground truth.}
We validate the effectiveness of our proposed goal inference algorithm, KD-BIL, by showing the correlation between the estimate posterior and the ground truth (in terms of the latent reward parameter $\mathcal{B}$) in Fig.5 on two maps, $3s\_vs\_5z$ and {\em corridor}. Each row in the heatmap shows the posterior distribution of one given reward parameter (which in the ideal case would concentrate on the diagonal line). The result shows that our proposed goal inference can accurately estimate the latent reward, even in complex tasks that involve a large number of blue/red agents and require advanced strategies (e.g, on the {\em corridor} map). The analysis of goal inference on other maps are provided in the appendix.

\textbf{Evaluating performance on different maps.} We deploy trained STUN agents alongside unknown agents on 6 different maps and repeat the experiments with several SOTA baselines for comparison: MAPPO~\cite{MAPPO}, COMA~\cite{COMA}, MAA2C, IPPO~\cite{IPPO}, IA2C~\cite{IA2C}. These baseline agents are trained using a multi-task learning approach by randomly sampling latent $\mathcal{B}$, so that they can collaborate with unknown agents of different behaviors/objectives.  
The results, as shown in Fig.~\ref{fig:smac}, demonstrate that the STUN's pre-training method can effectively converge and significantly improve teaming performance (up to 50\% on certain super-hard maps).
The settings are detailed in the Appendix.

\begin{figure}[tp]
\centering
  \subfigure[3s\_vs\_5z]{\includegraphics[width=0.40\textwidth]{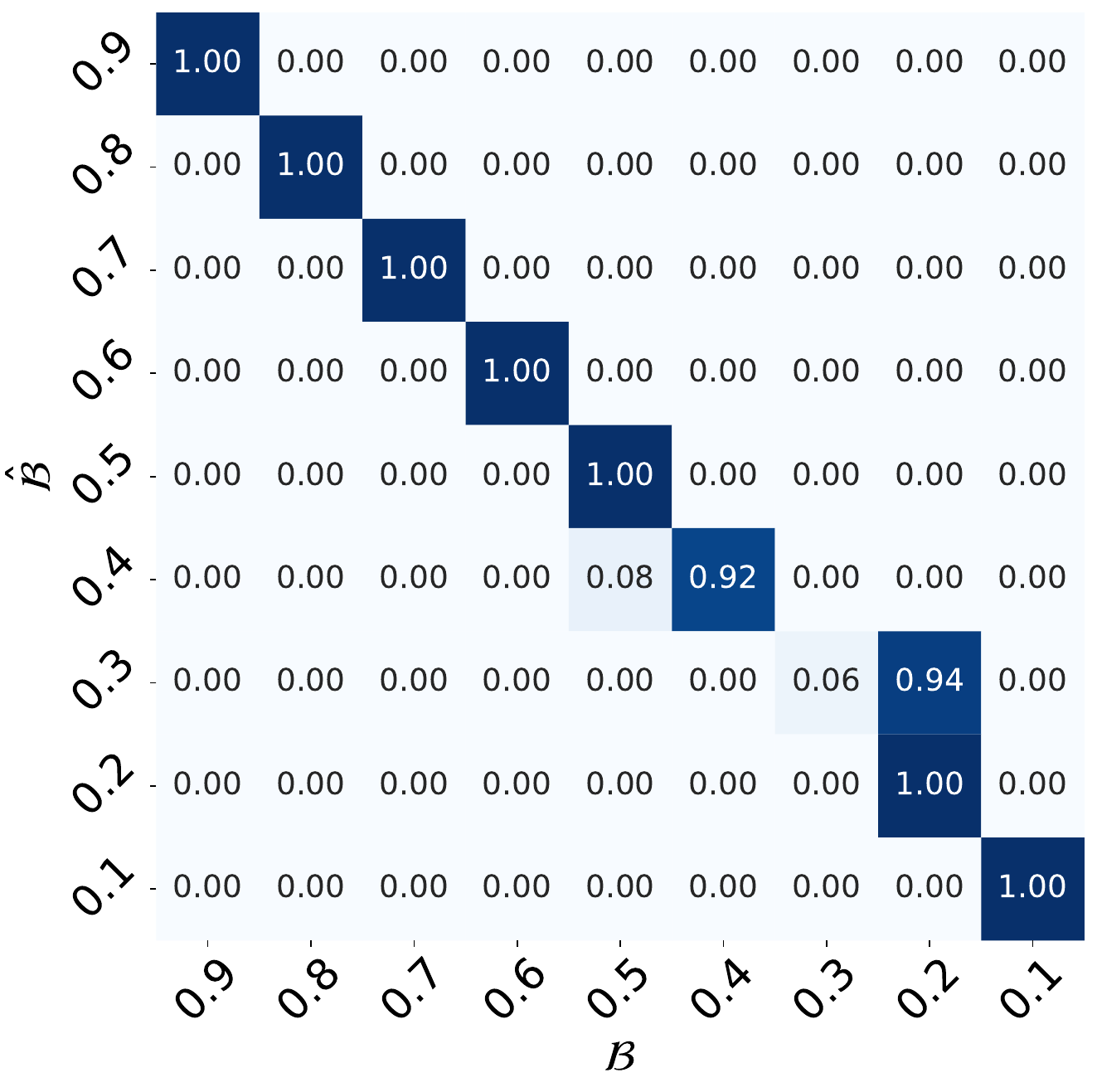}
  \label{fig:inverse1}
  }\hfill
  \subfigure[corridor]{\includegraphics[width=0.40\textwidth]{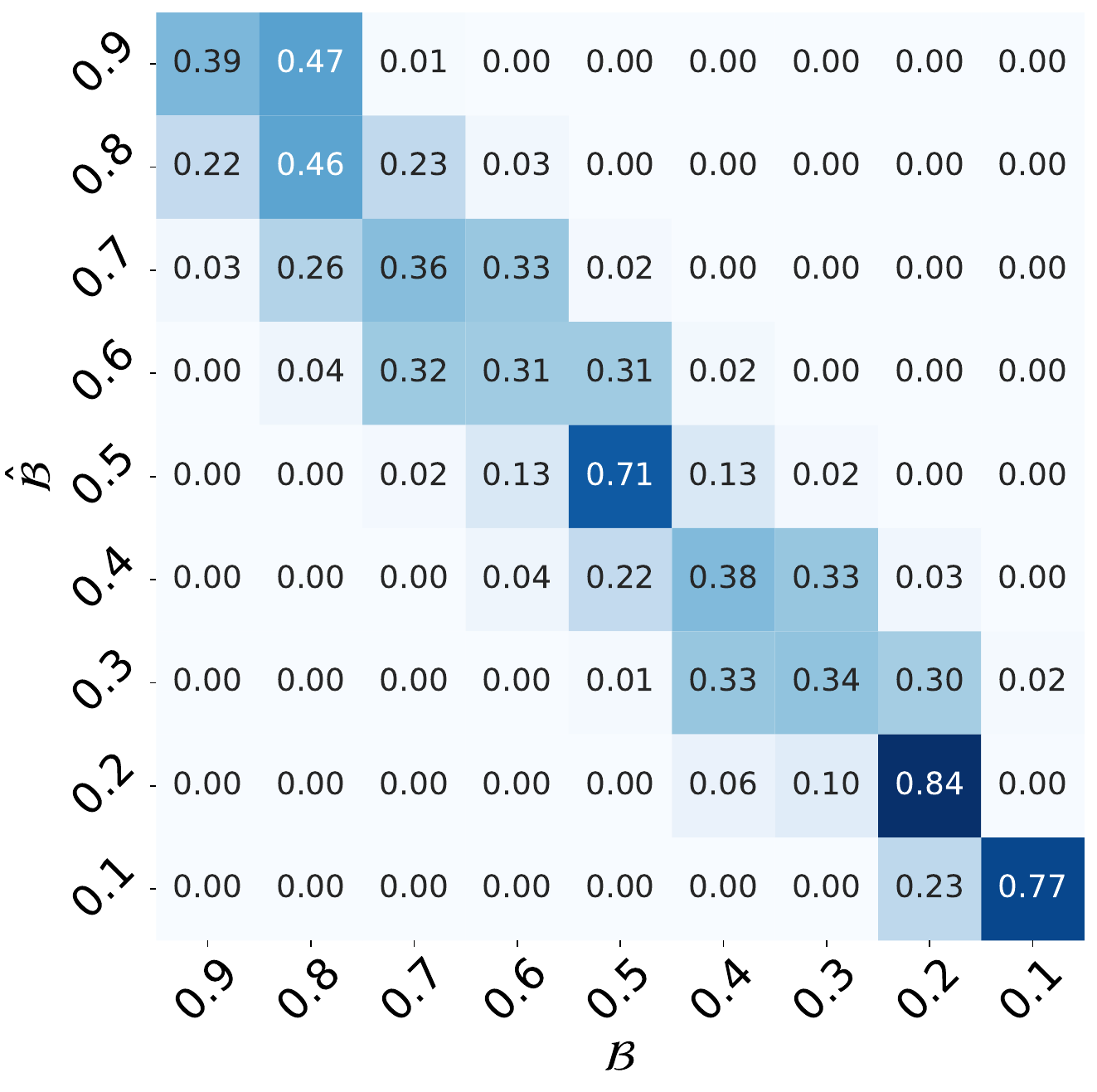}
  \label{fig:inverse2}
  }\hfill
  \caption{
An illustration of the correlation between the posterior estimate of reward parameters (shown in each row) using KD-BIL and the ground-truth reward parameters. Our proposed active goal inference can accurately infer the latent reward from observed unknown agent trajectories.}
\vspace{-0.2in}
  \label{fig:inverse}
\end{figure}

\textbf{Teaming performance with various unknown agents.}
We deploy the trained STUN agents alongside 8 different unknown agents on $3s\_v\_5z$ map and compare the teaming performance against two groups of baselines -- fixed-behavior agents and multi-task agents trained using different algorithms -- which are also deployed alongside the same unknown agents. Table~1 summarizes numerical results, with each row comparing the teaming performance of various collaborative agents alongside the same unknown agent. In particular, we calculate a normalized teaming score by assigning 100 points to the best-performing agent in each row 
and then taking the average over all 8 unknown agents. STUN agents achieve a normalized score of 99.2 out of a maximum of 100, with the best performance in nearly all scenarios and demonstrating robust teaming performance with a diverse range of unknown agents.

\section{Conclusions}

In this paper, we present STUN, a novel framework for enhancing AI and unknown-agent teaming in collaborative task environments. 
Considering unknown agents with latent rewards, we show that unbiased reward estimates are sufficient for optimal collaboration. By leveraging the KD-BIL algorithm for active goal inference and enabling a zero-shot policy adaptation, STUN can synergistically team up with unknown agents toward latent rewards. Evaluations using multi-agent environments including MPE and SMAC demonstrate robust teaming performance with diverse unknown agents and also agents with time-varying reward, outperforming a number of baselines. 
Synergistic teaming with unknown agents in non-stationary tasks or under restricted observations are avenues for future work.

\bibliography{main}
\bibliographystyle{unsrtnat}


\newpage
\appendix
\onecolumn

\section{Pseudo-code}
The pseudo-code of the proposed STUN framework is presented in Algorithm~\ref{alg:example}. 
The process can be divided into three stages: 1) Pre-train STUN agent, where STUN agents adaptable to various latent styles are trained. 2) Deploy STUN agent, where the STUN agent is deployed alongside other Unknown agents, and the trajectories of the Unknown agents are collected. 3) Inverse learning the latent reward, where the latent style of the Unknown agent is estimated. Furthermore, the STUN agent will compute $\hat{\mathcal{B}}$ and perform alongside the Unknown agents using $R_{\hat{\mathcal{B}}}$. Overall, the second and third stages can be combined into the Evaluation stage.

\begin{algorithm}
    \caption{Our Proposed STUN Framework}
    \label{alg:example}
\begin{algorithmic}
    \STATE  // Pre-train STUN agent
    \STATE Consider $n$ AI agents alongside $n_u$ surrogate agents.
    \FOR{$episode = 1$ {\bfseries to} $M$}
    \STATE Sample latent style $\mathcal{B}$ from a uniform distribution
    \FOR{$i=1$ {\bfseries to} ${n+n_u}$}
    \STATE Update policy with $\mathcal{B}$
    \ENDFOR 
    \ENDFOR
    \STATE  // Deploy STUN agent
    \STATE Replace surrogate agents with Unknown agents $\pi_u$
    \STATE STUN agents observe trajectories $\tau^u$ from unknown agents $\pi_u$ 
    \STATE  // Inverse learning the latent reward 
    \FOR{$i=1$ {\bfseries to} ${n}$ }
    \STATE Inverse learn $\hat{\mathcal{B}}$ as the latent style $\mathcal{B}$ of $\pi_u$
    \STATE Use $R_{\tilde{\mathcal{B}}}$ in the policy execution, where $\tilde{\mathcal{B}}$ is the mean of $\hat{\mathcal{B}}$.
    \ENDFOR
\end{algorithmic}
\end{algorithm}

\section{Mathematical Details}

\begin{proof}~\ref{lem:KD_BIRL}
According to Bayes' theorem, we can define each term as follows: the prior probability $p(R)$, the likelihood probability $\hat{p}_{m}(\{o_{i}^{u},a_{i}^{u}\}_{i=1}^{n}|R)$ and the marginal probability $\hat{p}_{m}(\{o_{i}^{u},a_{i}^{u}\}_{i=1}^{n})$:

\begin{equation}
\begin{aligned}
    \hat{p}_{m}(o_{i}^{u},a_{i}^{u}|R) & = \frac{\hat{p}_{m}(\{o_{i}^{u},a_{i}^{u}\}_{i=1}^{n}|R) p(R)}{\hat{p}_{m}(\{o_{i}^{u},a_{i}^{u}\}_{i=1}^{n})} \\
    & = \frac{\prod_{i=1}^{n}\hat{p}_{m}(o_{i}^{u},a_{i}^{u}|R) p(R)}{\hat{p}_{m}(\{o_{i}^{u},a_{i}^{u}\}_{i=1}^{n})} \\
    & \propto \prod_{i=1}^{n}\hat{p}_{m}(o_{i}^{u},a_{i}^{u}|R) p(R) \\
    & \propto p(R) \prod_{i=1}^{n}\sum\limits_{j=1}^{m}  \frac{K((o_{i}^{u},a_{i}^{u}),(o_{j},a_{j}))\cdot K_{R}(R,R_{j})}{\sum_{l=1}^{m}K_{R}(R,R_{l})}\\
    & \propto p(R) \prod\limits_{i=1}^{n} \sum\limits_{j=1}^{m} \frac{e^{-d_{s}((o_{i}^{u},a_{i}^{u}),(o_{j},a_{j}))^{2}/(2h)}e^{-d_{r}(R,R_{j})^{2}/(2h^{'})}}{\sum_{l=1}^{m}e^{-d_{r}(R,R_{l})^{2}/(2h^{'})}}
\end{aligned}
\end{equation}
\end{proof}

We begin by formulating an expression based on Bayes' theorem. Next, we incorporate  $\hat{p}_{m}(o_{i}^{u},a_{i}^{u}|R)$ into our analysis.

\begin{lemma}
\label{lem:random_process}
The random process $\{\Delta_{t}(x)\}$ defined as

$$\Delta_{t+1}(x) = (1-\alpha_{t}(x))\Delta_{t}(x) + \alpha_{t}(x)F_{t}(x)$$

converges to zero w.p.1 under the following assumptions:
\begin{itemize}
\item $0\leq \alpha_{t} \leq 1$, $\sum_{t}\alpha_{t}(x) = \infty$ and $ \sum_{t}\alpha_{t}^{2}(x) < \infty$
\item $\Vert \mathbb{E}[F_{t}(x)] \Vert_{q}\leq \gamma \Vert \Delta_{t} \Vert_{q}$, with $\gamma \leq 1$.
\item $Var(F_{t}(x))\leq C(1+\Vert \Delta_{t} \Vert_{q}^{2})$, for $C>0$.
\end{itemize}

Here, $\alpha_{t}(x)$ is allowed to depend on the past insofar as the above conditions remain valid.
\end{lemma}

\begin{proof}~\ref{lem:random_process}
See the lecture\cite{jaakkola1993convergence}
\end{proof}

\begin{proof}~\ref{thm:q_learning} we abbreviate $s_{t},s_{t+1},Q_{\mathcal{B},t},Q_{\mathcal{B},t+1}$ and $\alpha_{t}$ as $s,s^{'},Q_{\mathcal{B}},Q_{\mathcal{B}}^{'}$ and $\alpha$ respectively.

Subtract the optimal $Q^{*}_{\mathcal{B}}(s,a)$ from both sides in Eqn.~\ref{eql:q_learning}:

$$Q^{'}_{\mathcal{B}}(s,a) - Q^{*}_{\mathcal{B}}(s,a) = (1-\alpha)(Q_{\mathcal{B}}(s,a)-Q^{*}_{B}(s,a)) + \alpha[R_{\tilde{\mathcal{B}}}+\gamma \max\limits_{b\in \mathcal{A}}Q_{\mathcal{B}}(s^{\prime},b)-Q^{*}_{\mathcal{B}}(s,a)]$$

Let $\Delta_{t} = Q_{\mathcal{B}}(s,a)-Q^{*}_{\mathcal{B}}(s,a)$ and $F_{t}(s,a) = R_{\tilde{\mathcal{B}}} + \gamma \max\limits_{b\in \mathcal{A}}Q_{\mathcal{B}}(s^{\prime},b)-Q^{*}_{\mathcal{B}}(s,a)$

$$\Delta_{t+1}(s^{\prime},a) = (1-\alpha)\Delta_{t}(s,a) + \alpha F_{t}(s,a)$$

\begin{equation}
\begin{aligned}
    \mathbb{E}[F_{t}(s,a)] &= \sum\limits_{R_{\tilde{\mathcal{B}}}\in R} p(R_{\tilde{\mathcal{B}}}|s,a)  R_{\tilde{\mathcal{B}}} + \sum\limits_{s^{\prime}\in\mathcal{S}} p(s^{\prime}|s,a) \gamma \max\limits_{b\in\mathcal{A}} Q_{\mathcal{B}}(s^{\prime},b) - Q^{*}_{\mathcal{B}}(s,a) \\
    &= \sum\limits_{R_{\tilde{\mathcal{B}}}\in R} p(R_{\tilde{\mathcal{B}}}|s,a)  R_{\tilde{\mathcal{B}}} + \sum\limits_{s^{\prime}\in\mathcal{S}} p(s^{\prime}|s,a) \gamma \max\limits_{b\in\mathcal{A}} Q_{\mathcal{B}}(s^{\prime},b) - \sum\limits_{R_{\mathcal{B}}\in R} p(R_{\mathcal{B}}|s,a)  R_{\mathcal{B}} - \sum\limits_{s^{\prime}\in\mathcal{S}} p(s^{\prime}|s,a) \gamma \max\limits_{b\in\mathcal{A}} Q^{*}_{\mathcal{B}}(s^{\prime},b)\\
    &= \sum\limits_{R_{\tilde{\mathcal{B}}}\in R} p(R_{\tilde{\mathcal{B}}}|s,a)  R_{\tilde{\mathcal{B}}} - \sum\limits_{R_{\mathcal{B}}\in R} p(R_{\mathcal{B}}|s,a)  R_{\mathcal{B}} + \sum\limits_{s^{\prime}\in\mathcal{S}} p(s^{\prime}|s,a)  \gamma [\max\limits_{b\in\mathcal{A}} Q_{\mathcal{B}}(s^{\prime},b)  -  \max\limits_{b\in\mathcal{A}} Q^{*}_{\mathcal{B}}(s^{\prime},b)]\\
    &= \sum\limits_{s^{\prime}\in\mathcal{S}} p(s^{\prime}|s,a) \gamma [\max\limits_{b\in\mathcal{A}} Q_{\mathcal{B}}(s^{\prime},b)  -  \max\limits_{b\in\mathcal{A}} Q^{*}_{\mathcal{B}}(s^{\prime},b)]\\
    &\leq \gamma \sum\limits_{s^{\prime}\in\mathcal{S}} p(s^{\prime}|s,a) \max\limits_{b\in \mathcal{A},s^{\prime} \in \mathcal{S}} | Q_{\mathcal{B}}(s^{\prime},b) - Q^{*}_{\mathcal{B}}(s^{\prime},b) | \\
    &= \gamma \sum\limits_{s^{\prime}\in\mathcal{S}} p(s^{\prime}|s,a) \max\limits_{b\in \mathcal{A},s^{\prime} \in \mathcal{S}} \Vert Q_{\mathcal{B}} - Q^{*}_{\mathcal{B}} \Vert_{\infty} = \gamma \Vert Q_{\mathcal{B}} - Q^{*}_{\mathcal{B}} \Vert_{\infty} = \gamma \Vert \Delta_{t} \Vert_{\infty}
\end{aligned}
\end{equation}

Our initial step involves substituting the expression and simplifying it. We then apply Lemma~\ref{lem:linear_B} and Lemma~\ref{lem:noliner_B} to eliminate the terms $\sum\limits_{R_{\tilde{\mathcal{B}}}\in R} p(R_{\tilde{\mathcal{B}}}|s,a)  R_{\tilde{\mathcal{B}}} - \sum\limits_{R_{\mathcal{B}}\in R} p(R_{\mathcal{B}}|s,a)  R_{\mathcal{B}}$. The final step involves a scaling argument to complete the proof.

Finally,
\begin{equation}
\begin{aligned}
    Var[F_{t}(s,a)] &= \mathbb{E}[(F_{t}(s,a)-\mathbb{E}[F_{t}(s,a)])^{2}]\\
    &= \mathbb{E}\left[\left(R_{\tilde{\mathcal{B}}} + \gamma \max\limits_{b\in \mathcal{A}}Q_{\mathcal{B}}(s^{\prime},b) - [\sum\limits_{R_{\tilde{\mathcal{B}}}\in R} p(R_{\tilde{\mathcal{B}}}|s,a)  R_{\tilde{\mathcal{B}}} + \sum\limits_{s^{\prime}\in\mathcal{S}} p(s^{\prime}|s,a) \gamma \max\limits_{b\in\mathcal{A}} Q_{\mathcal{B}}(s^{\prime},b)]\right)^2\right]\\
    &= Var[R_{\tilde{\mathcal{B}}} + \gamma \max\limits_{b\in \mathcal{A}}Q_{\mathcal{B}}(s^{\prime},b)]
\end{aligned}
\end{equation}

Because $\hat{r}$ is bounded, it can be clearly verified that

$$Var[F_{t}(s,a)]\leq C(1+\Vert \Delta_{t} \Vert_{q}^{2})$$

for some constant $C$. Then, due to the Lemma~\ref{lem:random_process},$\Delta_{t}$ converges to zero w.p.1, i.e. $Q^{\prime}_{\mathcal{B}}(s,a)$ converges to $Q^{*}_{\mathcal{B}}(s,a)$ 
\end{proof}

\begin{proof}~\ref{lem:linear_B}
Based on the linear assumption of $R_{\mathcal{B}}(s,a)$, we can rewrite the expected reward as:
$$\mathbb{E}[R_{\tilde{\mathcal{B}}}(s,a)] =\mathbb{E}[\tilde{\mathcal{B}}^{T}R(s,a)] $$

Since $\tilde{\mathcal{B}} = \mathbb{E}[\mathcal{B}|\hat{\mathcal{B}}]$, we can replace $\tilde{\mathcal{B}}$ in the above equation with $\mathbb{E}[\mathcal{B}|\hat{\mathcal{B}}]$:

$$\mathbb{E}[R_{\tilde{\mathcal{B}}}(s,a)] =\mathbb{E}[\mathbb{E}[\mathcal{B}|\hat{\mathcal{B}}]^{T}R(s,a)] $$

Utilizing the Law of Iterated Expectations, we can combine the inner conditional expectation with the outer expectation:
$$\mathbb{E}[R_{\tilde{\mathcal{B}}}(s,a)] =\mathbb{E}[\mathcal{B}^{T}R(s,a)] $$

Since $\mathcal{B}$ is a fixed but unknown parameter, we can directly remove the expectation:
$$\mathbb{E}[R_{\tilde{\mathcal{B}}}(s,a)] =\mathcal{B}^{T}R(s,a) = R_{\mathcal{B}}(s,a)$$

Therefore, when the reward function is a linear function of latent parameters, $\tilde{\mathcal{B}} = \mathbb{E}[\mathcal{B}|\hat{\mathcal{B}}]$ indeed provides an unbiased estimate.
\end{proof}

\begin{proof}~\ref{lem:noliner_B}
We consider $R_{\tilde{\mathcal{B}}}$, which is $R(R^{-1}(\mathbb{E}[R_{\mathcal{B}}|\hat{\mathcal{B}}]))$. Since $R$ and $R^{-1}$ are inverse function of each other, this simplifies to $\mathbb{E}[R_{\mathcal{B}}|\hat{\mathcal{B}}]$.

Since $\tilde{\mathcal{B}}$ is defined based on $\mathbb{E}[R_{\mathcal{B}}|\hat{\mathcal{B}}]$, we can infer that $\mathbb{E}[R_{\tilde{\mathcal{B}}}] = \mathbb{E}[\mathbb{E}[R_{\mathcal{B}}|\hat{\mathcal{B}}]]$.

According to the Law of Iterated Expectations, $\mathbb{E}[\mathbb{E}[R_{\mathcal{B}}|\hat{\mathcal{B}}]] = \mathbb{E}[R_{\mathcal{B}}]$. 

Therefore, we have proven that $\mathbb{E}[R_{\tilde{\mathcal{B}}}] = \mathbb{E}[R_{\mathcal{B}}]$, which implies that $\tilde{\mathcal{B}}$ is an unbiased estimator of $\mathcal{B}$.

\end{proof}

\begin{proof}~\ref{lem:latentpolicy}
First, Let's start with the derivation of the state value function:

\begin{equation}
\begin{aligned}
    \nabla_{\theta}V_{\mathcal{B}}^{\pi_{\theta}}(s) &= \nabla_{\theta}(\sum\limits_{a\in \mathcal{A}}\pi_\theta(a|o,\mathcal{B}) Q_{\mathcal{B}}^{\pi_{\theta}}(s,a))\\
    &=\sum\limits_{a\in \mathcal{A}} (\nabla_{\theta}\pi_{\theta}(a|o,\mathcal{B})Q_{\mathcal{B}}^{\pi_{\theta}}(s,a) + \pi_{\theta}(a|o,\mathcal{B})\nabla_{\theta} Q_{\mathcal{B}}^{\pi_{\theta}}(s,a))\\
    &=\sum\limits_{a\in \mathcal{A}} (\nabla_{\theta}\pi_{\theta}(a|o,\mathcal{B})Q_{\mathcal{B}}^{\pi_{\theta}}(s,a) + \pi_{\theta}(a|o,\mathcal{B})\nabla_{\theta} \sum\limits_{s^{\prime},R_{\mathcal{B}}}p(s^{\prime},R_{\mathcal{B}}|s,a)(R_{\mathcal{B}}+\gamma V_{\mathcal{B}}^{\pi_{\theta}}(s^{\prime})))\\
    &=\sum\limits_{a\in \mathcal{A}} (\nabla_{\theta}\pi_{\theta}(a|o,\mathcal{B})Q_{\mathcal{B}}^{\pi_{\theta}}(s,a) + \gamma \pi_{\theta}(a|o,\mathcal{B}) \sum\limits_{s^{\prime},R_{\mathcal{B}}}p(s^{\prime},R_{\mathcal{B}}|s,a)\nabla_{\theta} V_{\mathcal{B}}^{\pi_{\theta}}(s^{\prime}))\\
    &=\sum\limits_{a\in \mathcal{A}} (\nabla_{\theta}\pi_{\theta}(a|o,\mathcal{B})Q_{\mathcal{B}}^{\pi_{\theta}}(s,a) + \gamma \pi_{\theta}(a|o,\mathcal{B}) \sum\limits_{s^{\prime}}p(s^{\prime}|s)\nabla_{\theta} V_{\mathcal{B}}^{\pi_{\theta}}(s^{\prime}))
\end{aligned}
\end{equation}

To simplify the representation, let $\phi(s) = \sum_{a\in \mathcal{A}}\nabla_{\theta}\pi_{\theta}(a|o,\mathcal{B})Q_{\mathcal{B}}^{\pi_{\theta}}(s,a)$, and define $d^{\pi_{\theta}}(s\rightarrow x,k)$ as the probability that a strategy arrives at a state after starting a step from the state.

\begin{equation}
\begin{aligned}
    \nabla_{\theta}V_{\mathcal{B}}^{\pi_{\theta}}(s) &= \phi(s) + \gamma \sum\limits_{a\in \mathcal{A}}\pi_{\theta}(a|o,\mathcal{B})\sum\limits_{s^{'}\in \mathcal{S}}p(s^{'}|s,a)\nabla_{\theta}V_{\mathcal{B}}^{\pi_{\theta}}(s^{'})\\
    &= \phi(s) + \gamma \sum\limits_{a\in \mathcal{A}}\sum\limits_{s^{'}\in \mathcal{S}}\pi_{\theta}(a|o,\mathcal{B})p(s^{'}|s,a)\nabla_{\theta}V_{\mathcal{B}}^{\pi_{\theta}}(s^{'})\\
    &= \phi(s) + \gamma \sum\limits_{s^{'}\in \mathcal{S}} d^{\pi_{\theta}}(s\rightarrow s^{'},1)\nabla_{\theta}V_{\mathcal{B}}^{\pi_{\theta}}(s^{'})\\
    &= \phi(s) + \gamma \sum\limits_{s^{'}\in \mathcal{S}} d^{\pi_{\theta}}(s\rightarrow s^{'},1)[\phi(s^{'}) + \gamma \sum\limits_{s^{''}\in \mathcal{S}} d^{\pi_{\theta}}(s^{'}\rightarrow s^{''},1)\nabla_{\theta}V_{\mathcal{B}}^{\pi_{\theta}}(s^{''})]\\
    &= \phi(s) + \gamma \sum\limits_{s^{'}\in \mathcal{S}} d^{\pi_{\theta}}(s\rightarrow s^{'},1)\phi(s^{'}) + \gamma^{2} \sum\limits_{s^{''}\in \mathcal{S}} d^{\pi_{\theta}}(s\rightarrow s^{''},2)\nabla_{\theta}V_{\mathcal{B}}^{\pi_{\theta}}(s^{''})]\\
    & = ...\\
    & = \sum\limits_{x\in \mathcal{S}}\sum\limits_{k=0}^{\infty} \gamma^{k} d^{\pi_{\theta}}(s\rightarrow x,k) \phi(x)
\end{aligned}
\end{equation}

Define $\eta(s)=\mathbb{E}_{s_{0}\sim\mu, s, a}[\sum_{k=0}^{\infty}\gamma^{k}d^{\pi_{\theta}}(s_{0} \rightarrow s,k)]$. With this, we return to the objective function:"

\begin{equation}
\begin{aligned}
    J_{\pi,\mathcal{B}}(\theta)& = \mathbb{E}_{s_{0}\sim\mu, s, a} [V^{\pi_{\theta}}_{\mathcal{B}}(s_{0})]\\
    & = \sum\limits_{s\in\mathcal{S}}\mathbb{E}_{s_{0}\sim\mu, s, a}[\sum\limits_{k=0}^{\infty} \gamma^{k} d^{\pi_{\theta}}(s\rightarrow x,k)]\phi(s)\\
    & = \sum\limits_{s\in\mathcal{S}}\eta(s)\phi(s)\\
    & = (\sum\limits_{s\in\mathcal{S}}\eta(s))\sum\limits_{s\in\mathcal{S}}\frac{\eta(s)}{\sum\limits_{s\in\mathcal{S}}\eta(s)}\phi(s)\\
    & \propto \frac{\eta(s)}{\sum\limits_{s\in\mathcal{S}}\eta(s)}\phi(s)\\
    & = \sum\limits_{s\in\mathcal{S}}V_{\mathcal{B}}^{\pi_{\theta}}(s)\sum\limits_{a\in \mathcal{A}}Q^{\pi_{\theta}}_{\mathcal{B}}(s,a)\nabla_{\theta}\pi_{\theta}(a|o,\mathcal{B})\\
    & = \mathbb{E}_{\pi_{\theta}}[Q^{\pi_{\theta}}_{\mathcal{B}}(s,a)\nabla_{\theta}\log \pi_{\theta}(a|o,{B})]
\end{aligned}
\end{equation}

\end{proof}

\section{Environment Details}
\subsection{MPE}

\subsubsection{Implementation details and Hyper-parameters}
We modified the Predator-prey task in MPE to illustrate different behaviors of various latent styles. Specifically, several agents must navigate a $300\times 300$ $2$-dimensional Euclidean space, where the map size is customizable. Each agent and adversary can perform one of five actions: remain stationary or move in one of four directions. Each agent has a resource radius, $r_{source}$. This radius implies that when another agent enters it, a reward of $-1$ is received, which is referred to as the greedy reward. Every adversary has a predation radius, $r_{adv}$. This radius means that when another agent enters it, the agent receives a reward of $-1$, known as the Safety reward. From this, we can distinguish two different latent styles. Of course, the reward values can also be set differently, such as based on the distance between two units. Cost reward and Preference reward are additional supplements for more latent behavioral styles. The Cost reward occurs when an agent's chosen action is to move, thereby consuming energy and receiving a reward of $-1$. The Preference reward, on the other hand, reflects a preference for different values when the previously obtained reward is not $-1$.

The overall execution process begins at the start of each epoch. $N$ agents and $M$ adversaries are randomly placed on the plane. The strategy for adversaries is always to choose the nearest agent, and different adversaries will select different agents. Then, the process runs for 100 steps to count all reward outcomes, which are returned as the environment's output.

\subsubsection{Interpretability}
In the reconfigured environment we set up, the relationship between predators and prey from the natural world is authentically simulated. When observing lions and herds of cattle on the grassland, we found that the cattle would cluster together to flee from predators. We wanted to replicate this process. We discovered that when only influenced by the Safety reward, all agents would always gather together to escape from the pursuit of adversaries. This is similar to the patterns observed in nature (although in nature, there is behavior to protect young animals in the middle of the herd). To exhibit different styles, we introduced the Greedy reward. This means that when an agent is moving away from an adversary, the agent also exhibits a certain level of selfishness, desiring to possess more resources. This implies that when only the Greedy reward is in effect, agents, in pursuit of maximum resources, will be evenly distributed across the entire two-dimensional space. By combining these two types of rewards in varying magnitudes, different contraction radii emerge as they move away from adversaries. We have showcased the behaviors we observed in Fig.~\ref{fig:render}.
We can also characterize different types of agents through the Preference reward. If an agent receives a larger reward when it is closer to other agents, this indicates that the agent has a lesser impact on other agents. In the natural world, this might correspond to younger prey, which possess fewer resources. As a result, such an agent will always be positioned in the middle of other agents.

\begin{figure*}[h]
\label{fig:render}
\centering
  \subfigure[Type of units]{
  \label{fig:render1}
  \includegraphics[width=0.15\textwidth]{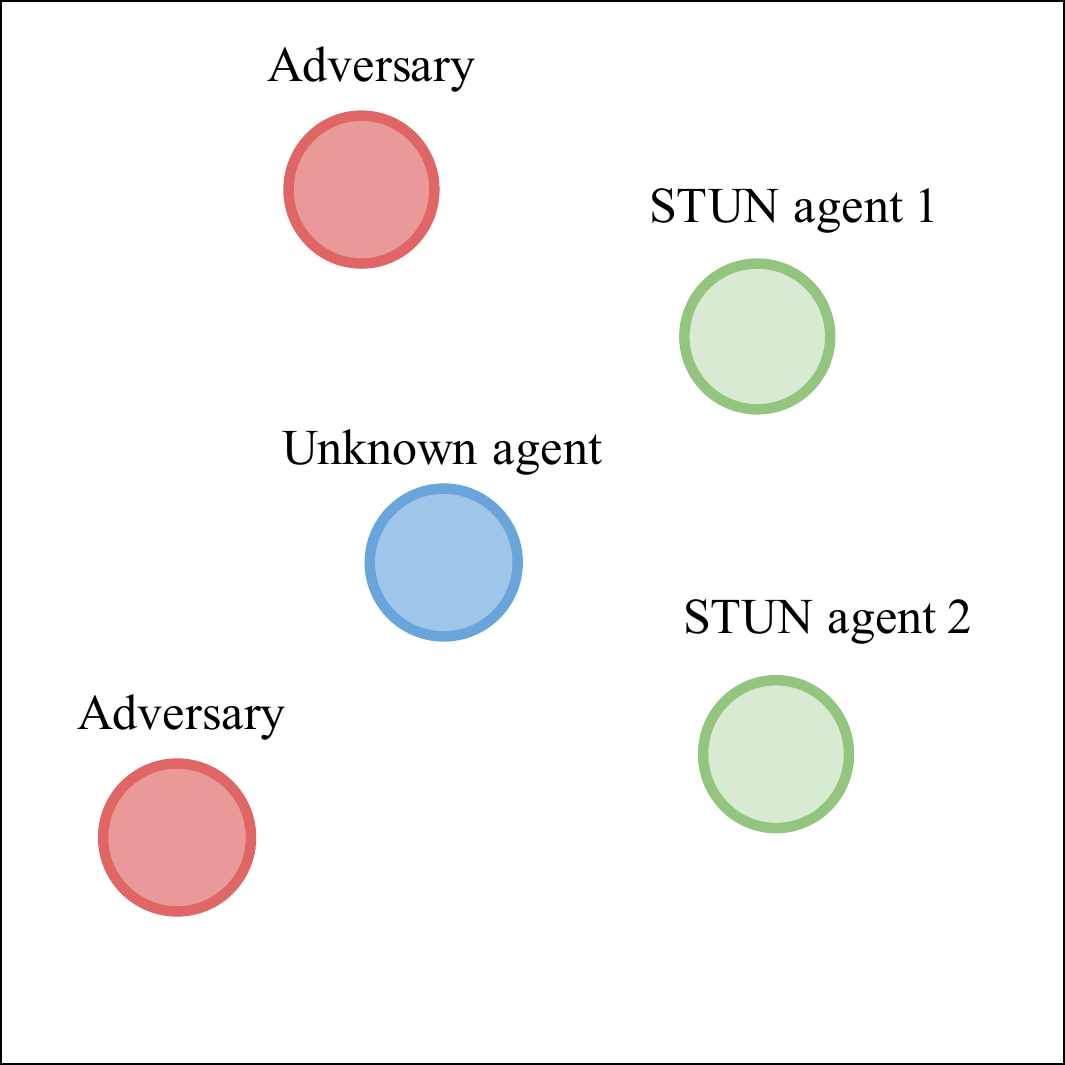}}\hfill
  \subfigure[Greedy]{
  \label{fig:render2}
  \includegraphics[width=0.15\textwidth]{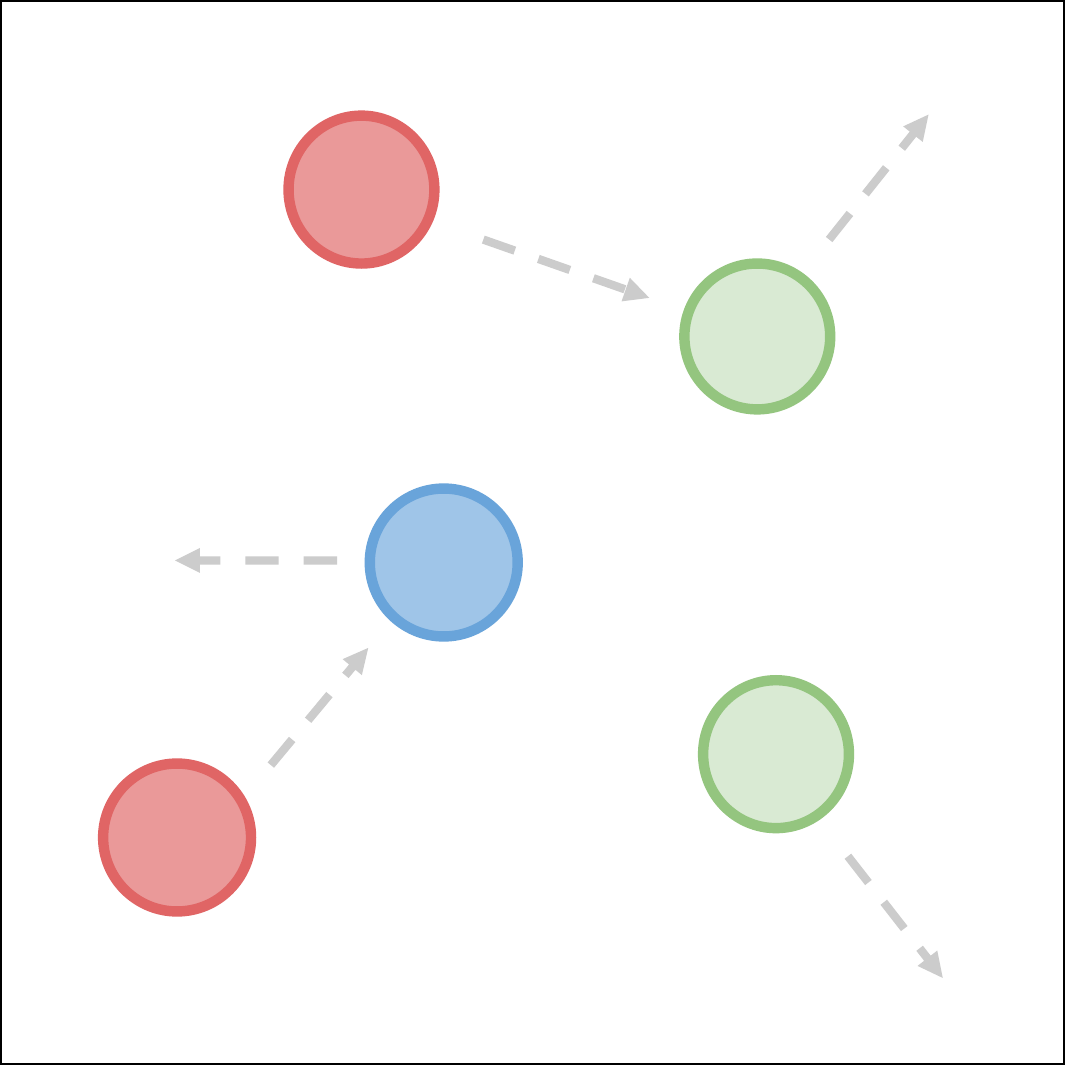}}\hfill
  \subfigure[Safety]{
  \label{fig:render3}
  \includegraphics[width=0.15\textwidth]{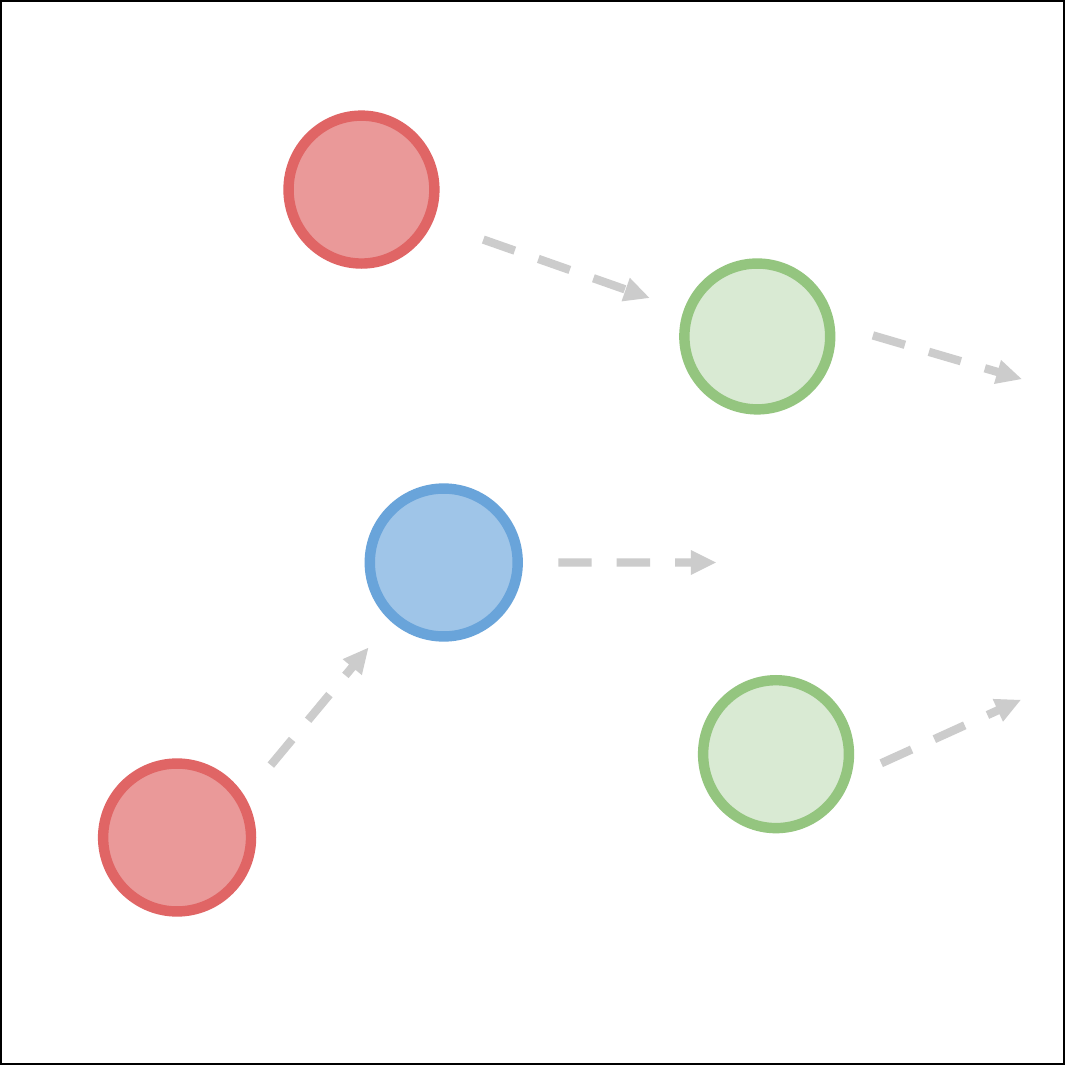}}\hfill
  \subfigure[Greedy]{
  \label{fig:render4}
  \includegraphics[width=0.15\textwidth]{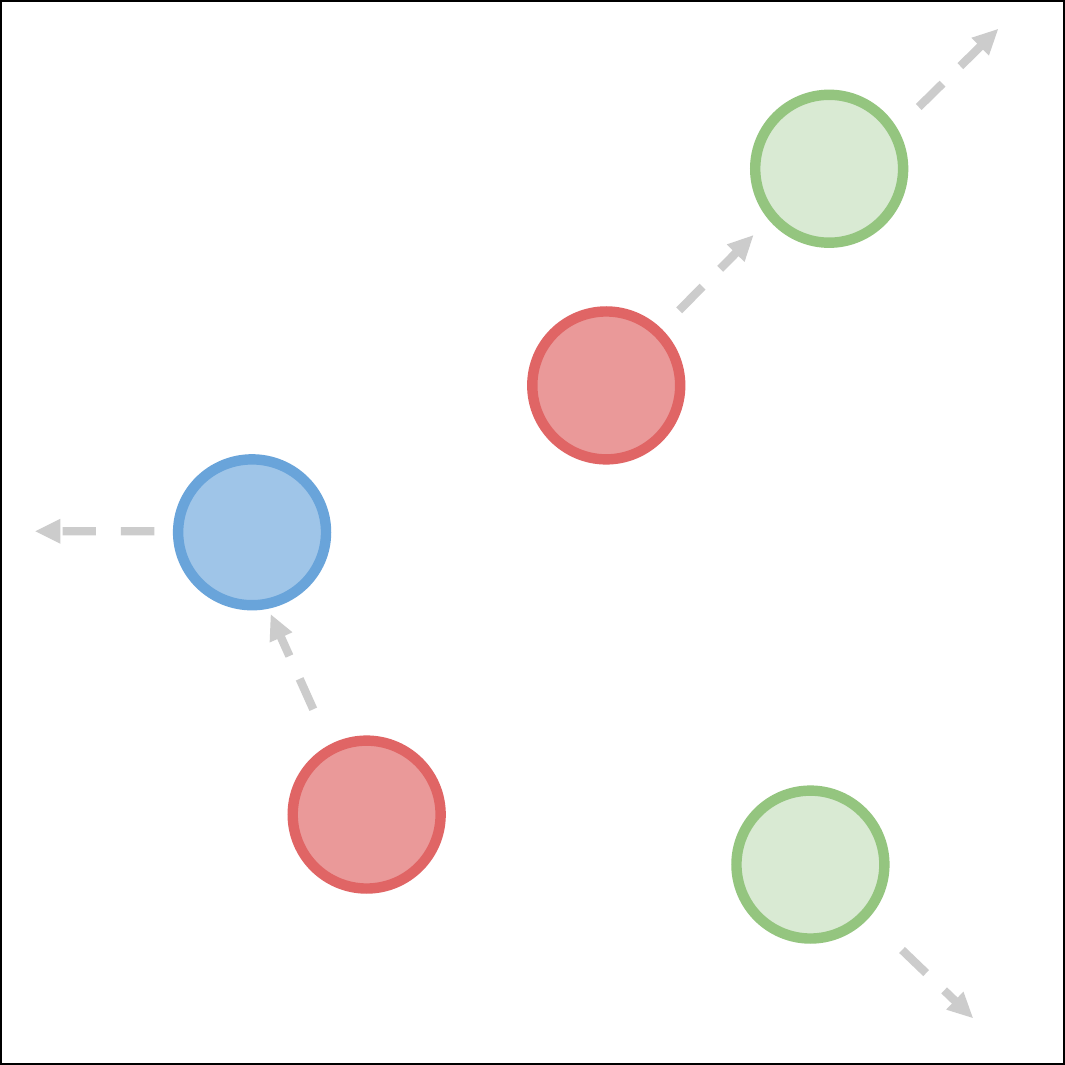}}\hfill
  \subfigure[Balance]{
  \label{fig:render5}
  \includegraphics[width=0.15\textwidth]{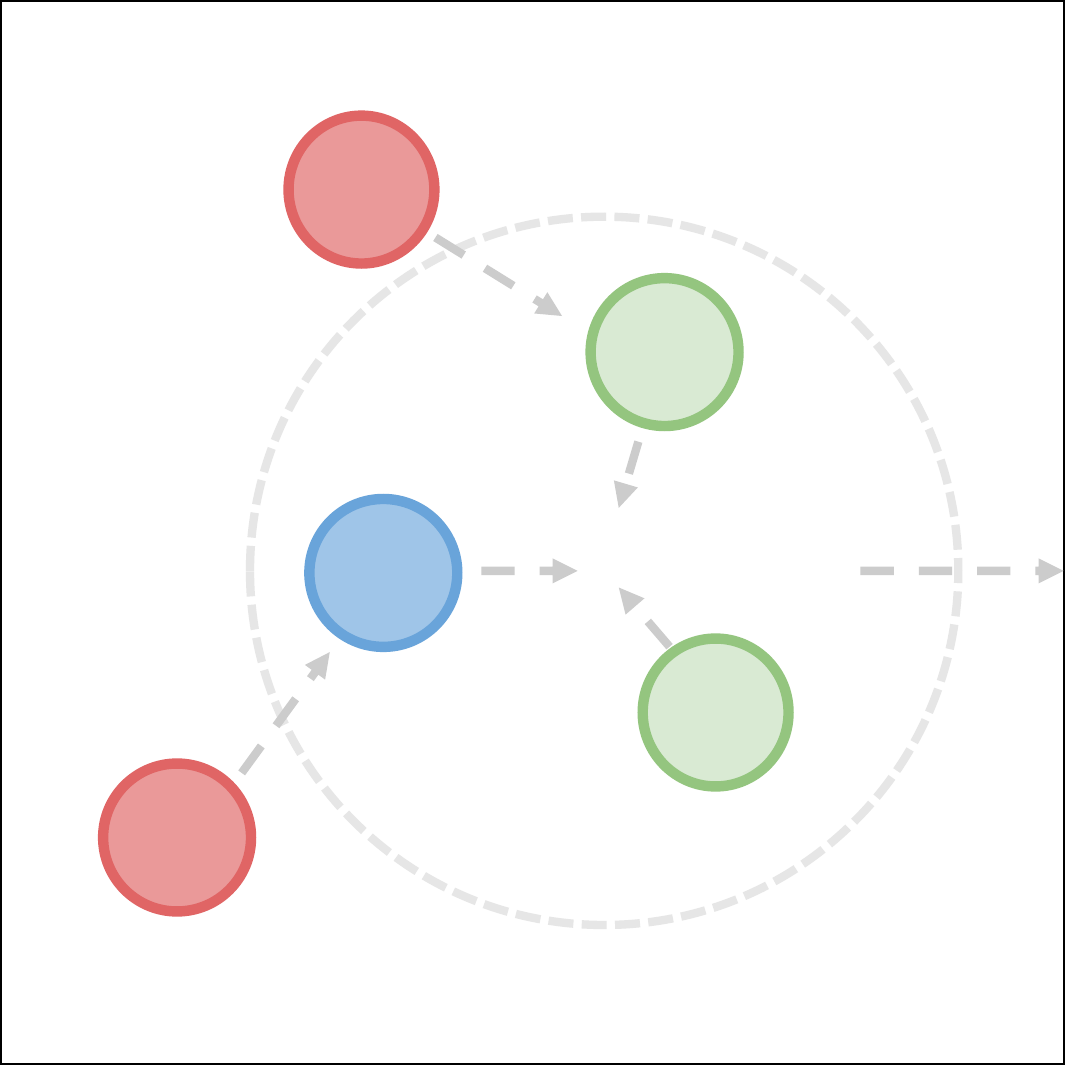}}\hfill
  \subfigure[Safety]{
  \label{fig:render6}
  \includegraphics[width=0.15\textwidth]{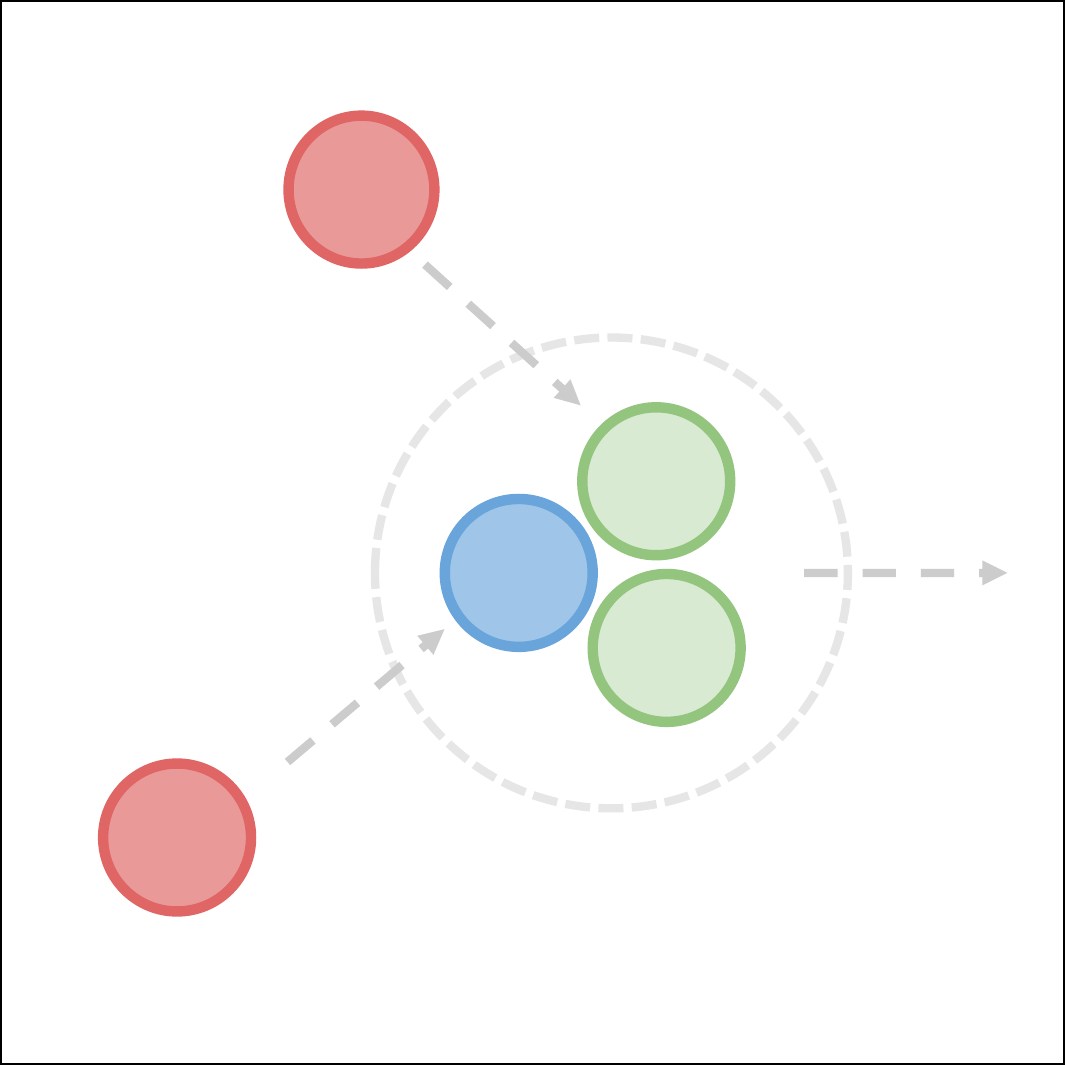}}\hfill
  \caption{
  This is an illustrative diagram explaining different parameter styles $\mathcal{B}$. Here, Fig.~\ref{fig:render1} displays the types of units in the diagram. Fig.~\ref{fig:render2} refers to the scenario where if the agent's style is completely greedy, then all agents' actions would be as depicted: adversaries move towards the nearest agent, while agents ignore adversaries and repel each other, eventually achieving a uniform distribution in space. If the agent's style is completely safe, as shown in Fig.~\ref{fig:render3}, then all agents will stay away from adversaries and cluster together at a distance. In figures Fig.~\ref{fig:render4} to~\ref{fig:render6}, the transition from a greedy to a safe agent style is shown, starting from distancing from each other to clustering together to escape the adversary, with the distance between clustering agents being greater for more greedy agents and smaller for safer ones.
  }
\end{figure*}

To better explore the relationship between different values of $N$ and $M$, we conducted the following experiment: We set up scenarios with varying quantities of $N$ and $M$. This illustrates both the impact of different numbers of agents on behavior and the influence of the relative values of different rewards. By observing the plots in Fig.~\ref{fig:2reward}, we found that the trend obtained between the Greedy reward and Safety reward for all agents is the same and that different numbers of agents lead to variations in the magnitude of rewards obtained.
\begin{figure*}[h]
\label{fig:2reward}
\centering
  
  \subfigure[N=3,M=1]{\includegraphics[width=0.33\textwidth]{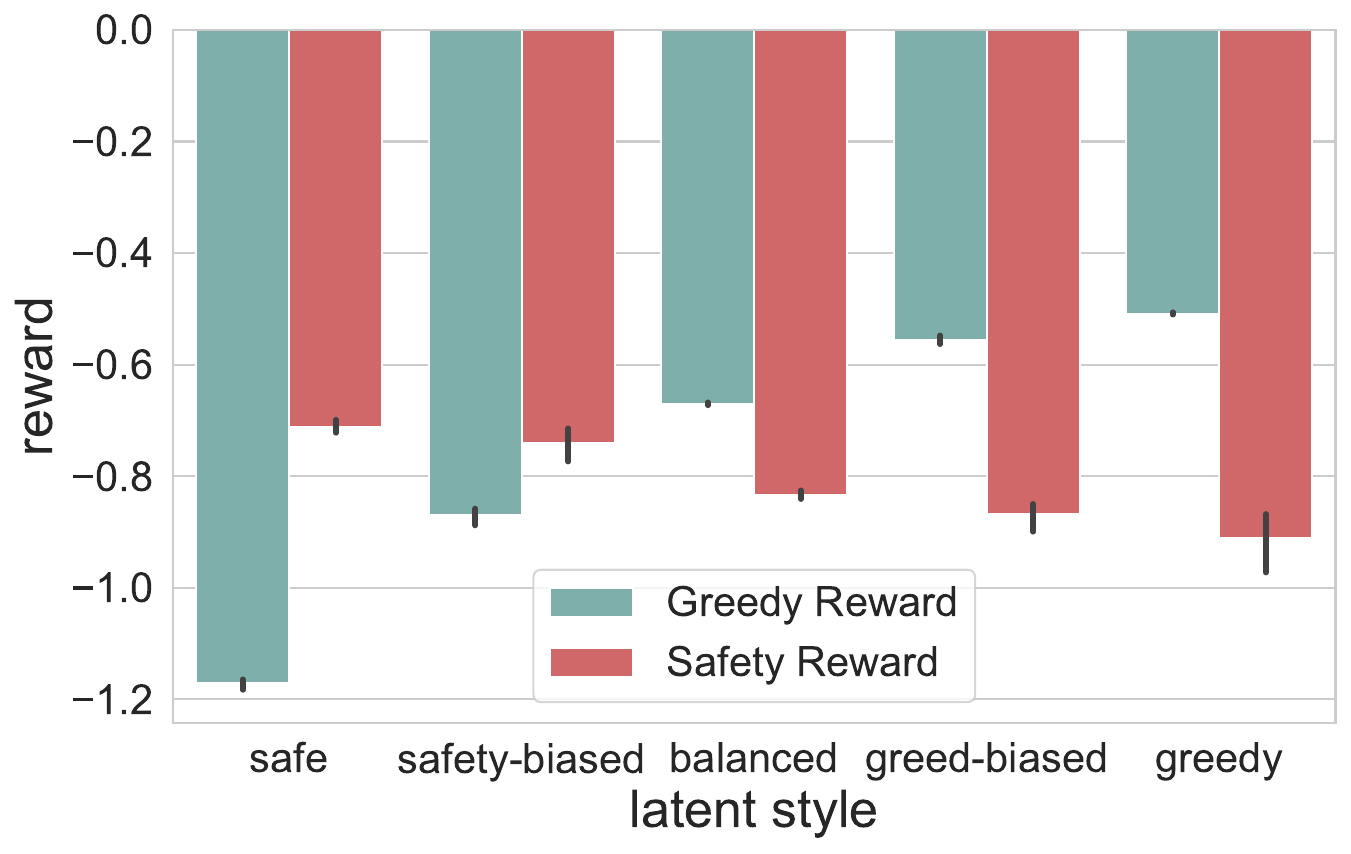}}\hfill
  \subfigure[N=3,M=2]{\includegraphics[width=0.33\textwidth]{data/mpe/reward_component_3_2.pdf}}\hfill
  \subfigure[N=3,M=4]{\includegraphics[width=0.33\textwidth]{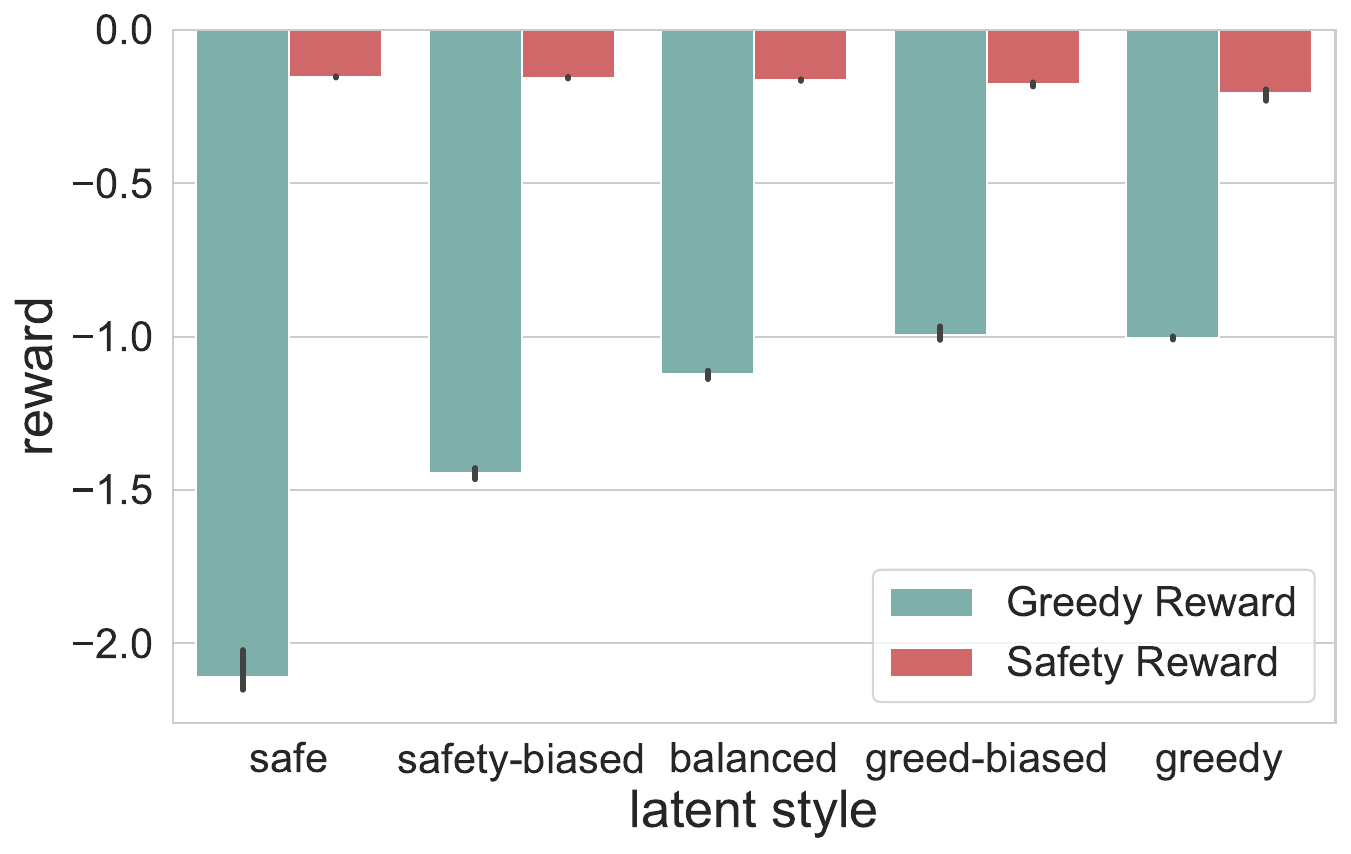}}\hfill
  \\
  \subfigure[N=5,M=2]{\includegraphics[width=0.33\textwidth]{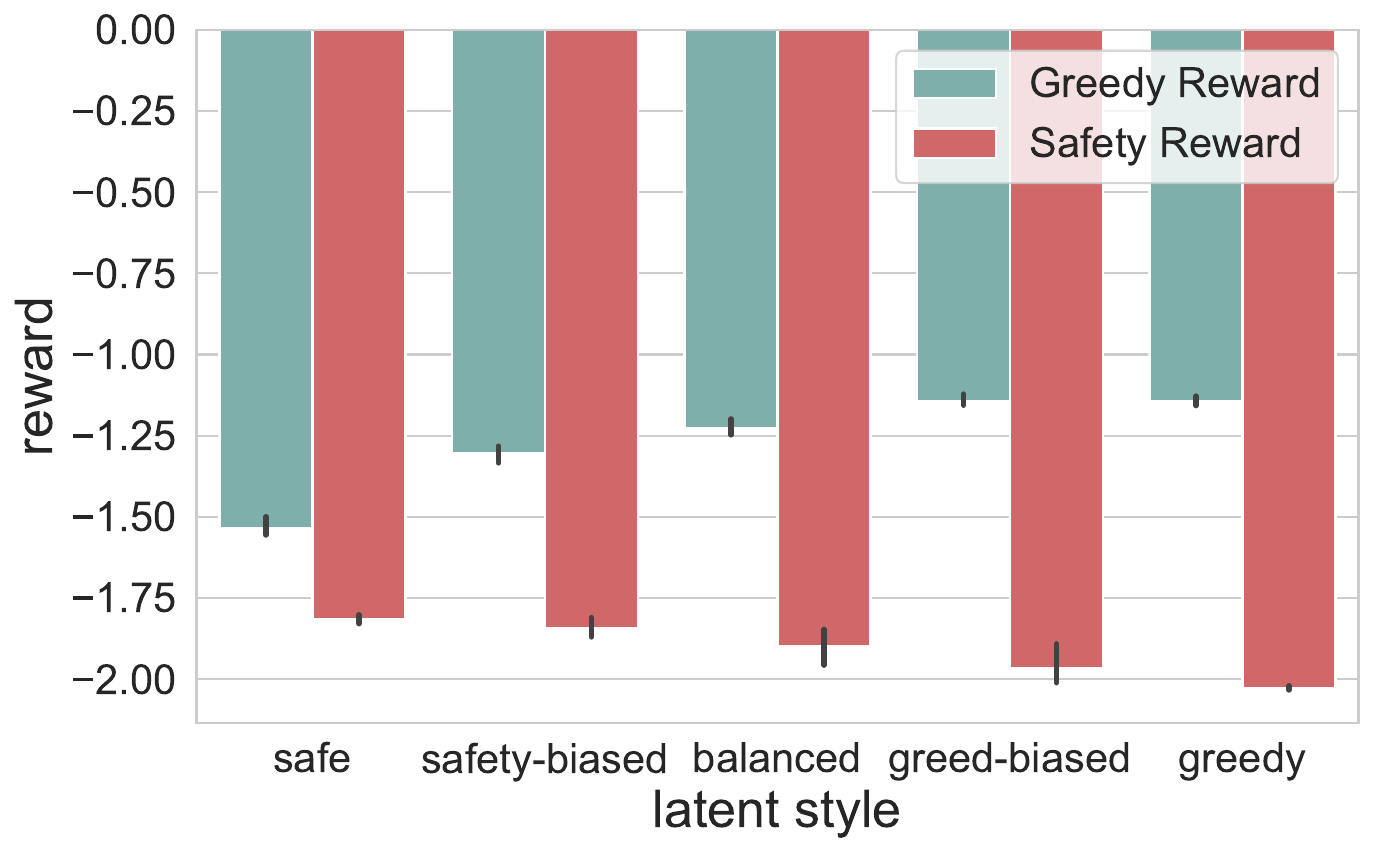}}\hfill
  \subfigure[N=5,M=4]{\includegraphics[width=0.33\textwidth]{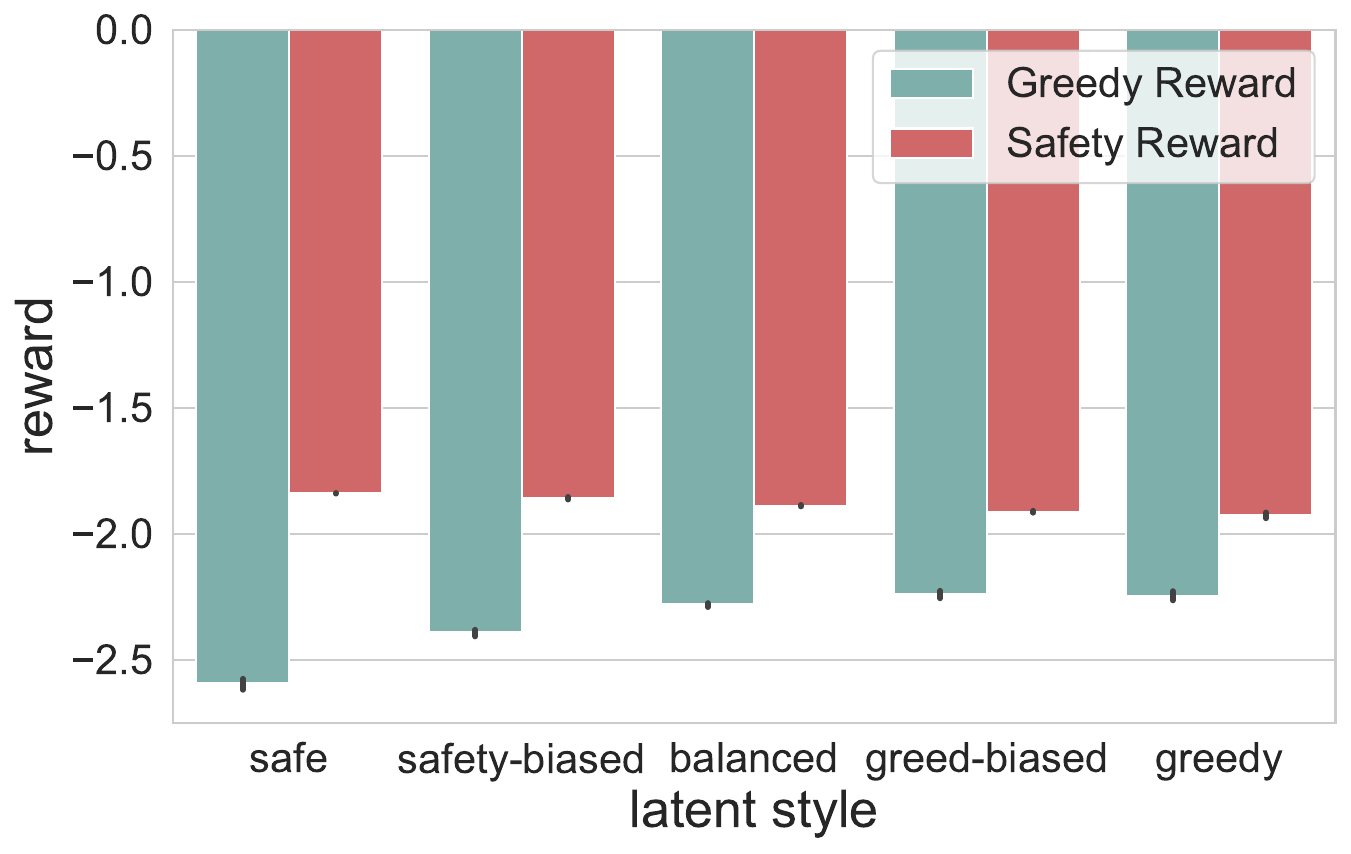}}\hfill
  \subfigure[N=5,M=7]{\includegraphics[width=0.33\textwidth]{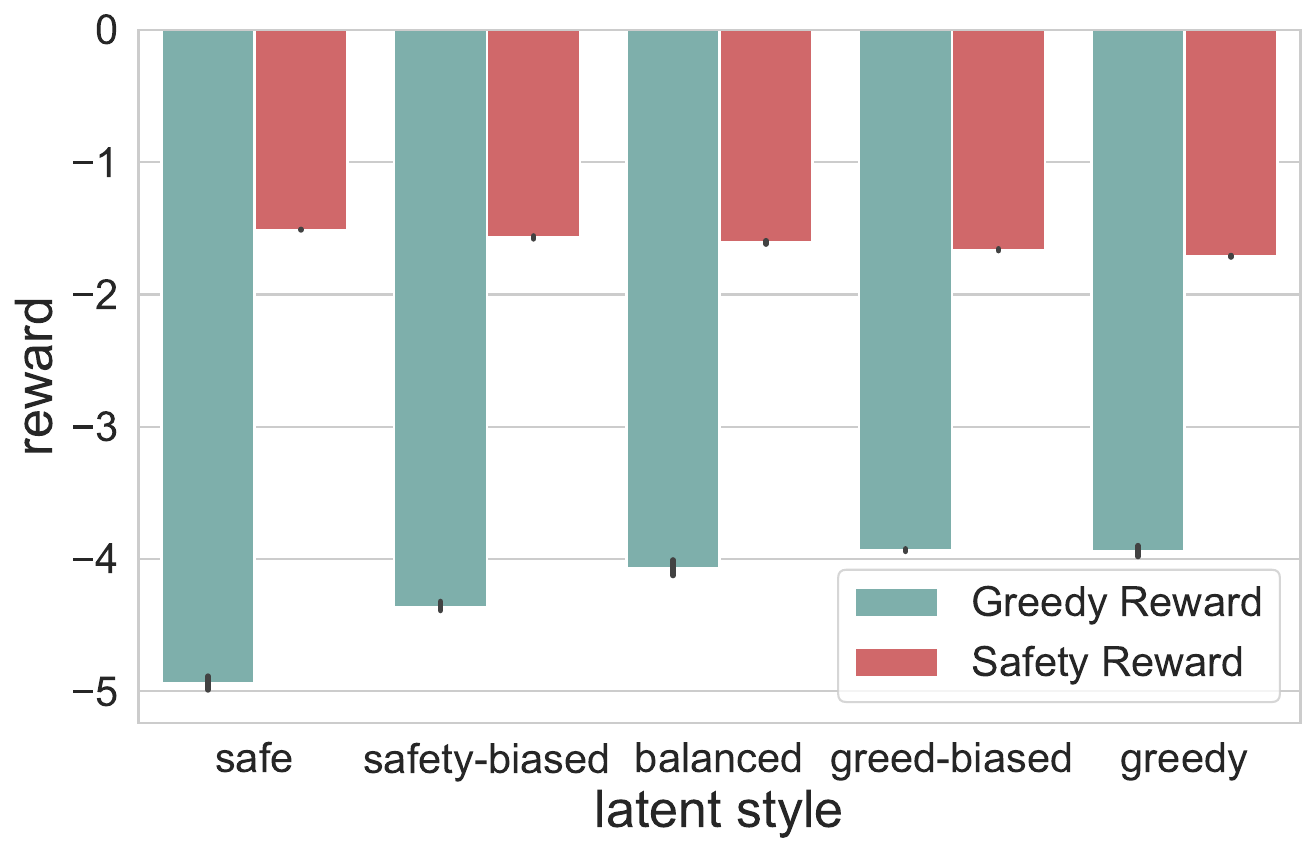}}\hfill\vspace{-1.ex}
  \caption{Different quantities of Greed reward and Safety reward components, denoted by $N$ and $M$ respectively.} 
\end{figure*}

\subsection{Collaborating with dynamic agents with varying}
We adopted six combinations of $N$ and $M$ for testing. We divided the results into two groups based on the different values of $N$. In each group's test force, we cooperated with Surrogate agents and STUN agents. The Surrogate agent undergoes a $\mathcal{B}$ change every 20 epochs, and each epoch consists of 25 steps of training. Each STUN agent maintains a queue of 300 lengths, storing the observed trajectories of the Surrogate agent. At the beginning of each epoch, each STUN updates $\hat{\mathcal{B}}$ using the data from the queue and cooperates with the Surrogate agent using $\hat{\mathcal{B}}$ to obtain the test reward. From this figure, we can see that when the $\mathcal{B}$ of the Surrogate agent changes, the system's reward decreases. This is because the STUN agent cannot cooperate well with the Surrogate agent. However, as the data in the queue is updated, the STUN can estimate $\hat{\mathcal{B}}$ well, and the system's reward will increase and stabilize within a range.
\begin{figure*}[h]
\centering
  
  \subfigure[N=3]{\includegraphics[width=0.5\textwidth]{data/mpe/dynamic3.pdf}}\hfill
  \subfigure[N=5]{\includegraphics[width=0.5\textwidth]{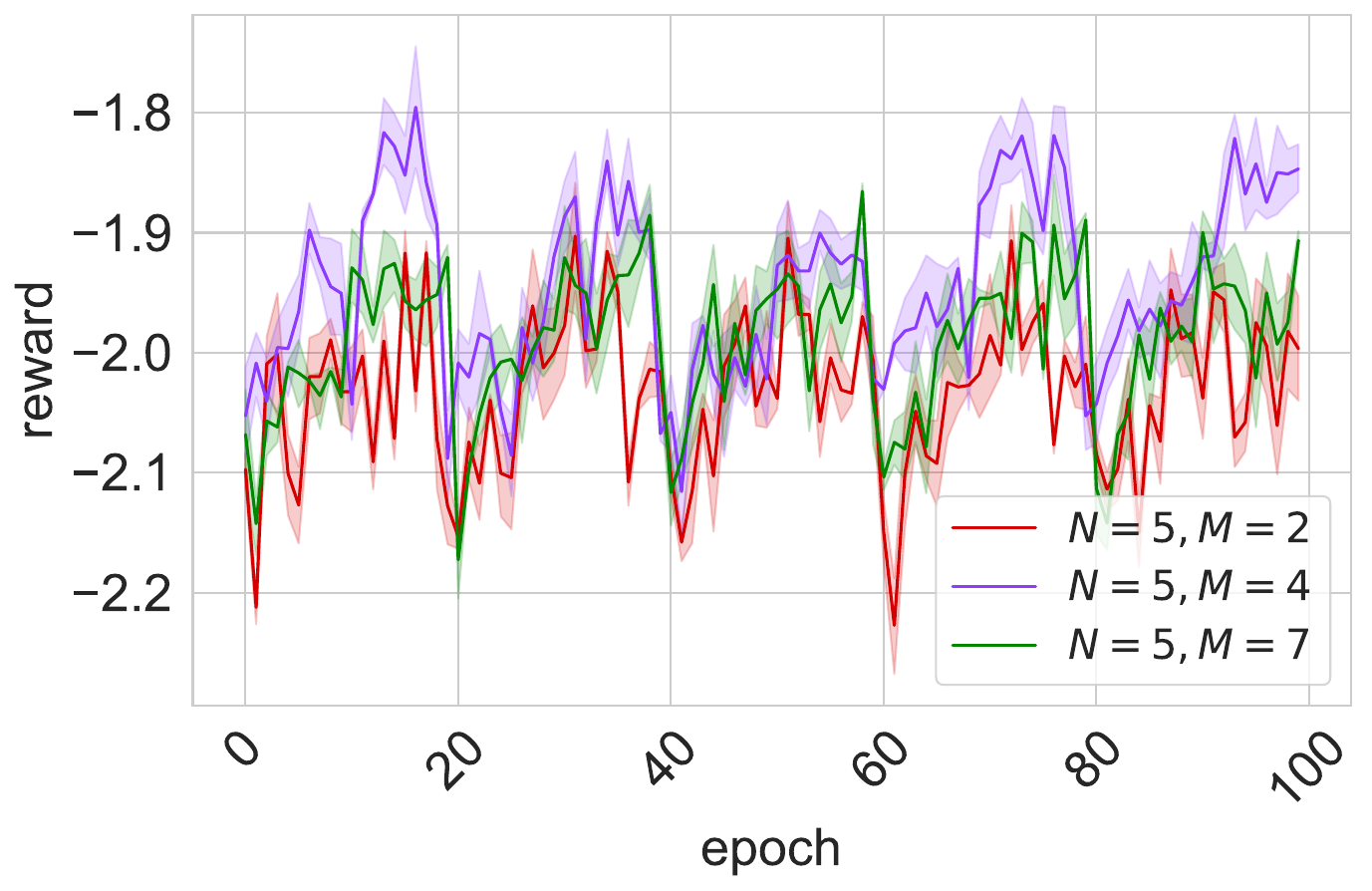}}\hfill\vspace{-2.5ex}
  \caption{Every 20 epochs, the Surrogate agent changes $\mathcal{B}$. Meanwhile, each STUN agent continuously collects the trajectories of the Surrogate agent. With the gradual collection of new trajectories from the Surrogate agent, the STUN agent can better update the estimated $\mathcal{B}$, thereby achieving an increase in reward.}
\end{figure*}

\subsection{Ablation study}
To validate that our algorithm can satisfy high-dimensional, linear, and nonlinear reward functions, we conducted Pre-training experiments. Moreover, to verify the necessity of the Surrogate during the Pre-training phase, we replaced it with both a fixed-style Unknown agent and a random-style Unknown agent for training together, resulting in the following set of plots in Fig.~\ref{fig:ablation_all}. Each set of results represents a different combination of $N$ and $M$. From observations, it can be found that the STUN Pre-training framework can converge under high-dimensional, linear, and nonlinear reward function scenarios. Furthermore, the Surrogate agent can effectively assist in convergence, whereas training with either a fixed-style or a random-style Unknown agent does not achieve satisfactory results.

\begin{figure*}[h]
\label{fig:ablation_all}
\centering
  
  \subfigure[N=3,M=1]{\includegraphics[width=0.33\textwidth]{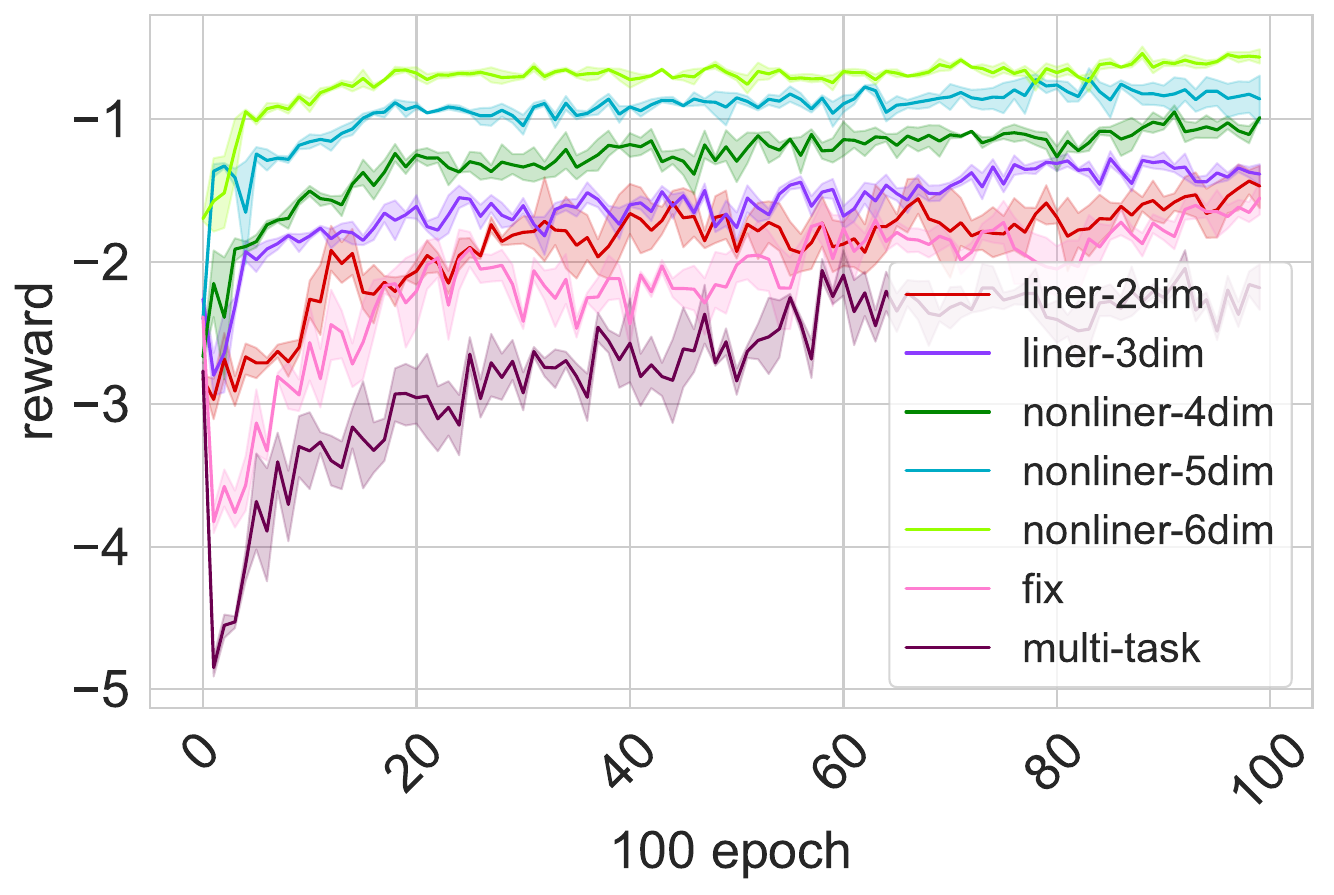}}\hfill
  \subfigure[N=3,M=2]{\includegraphics[width=0.33\textwidth]{data/mpe/3_2.pdf}}\hfill
  \subfigure[N=3,M=4]{\includegraphics[width=0.33\textwidth]{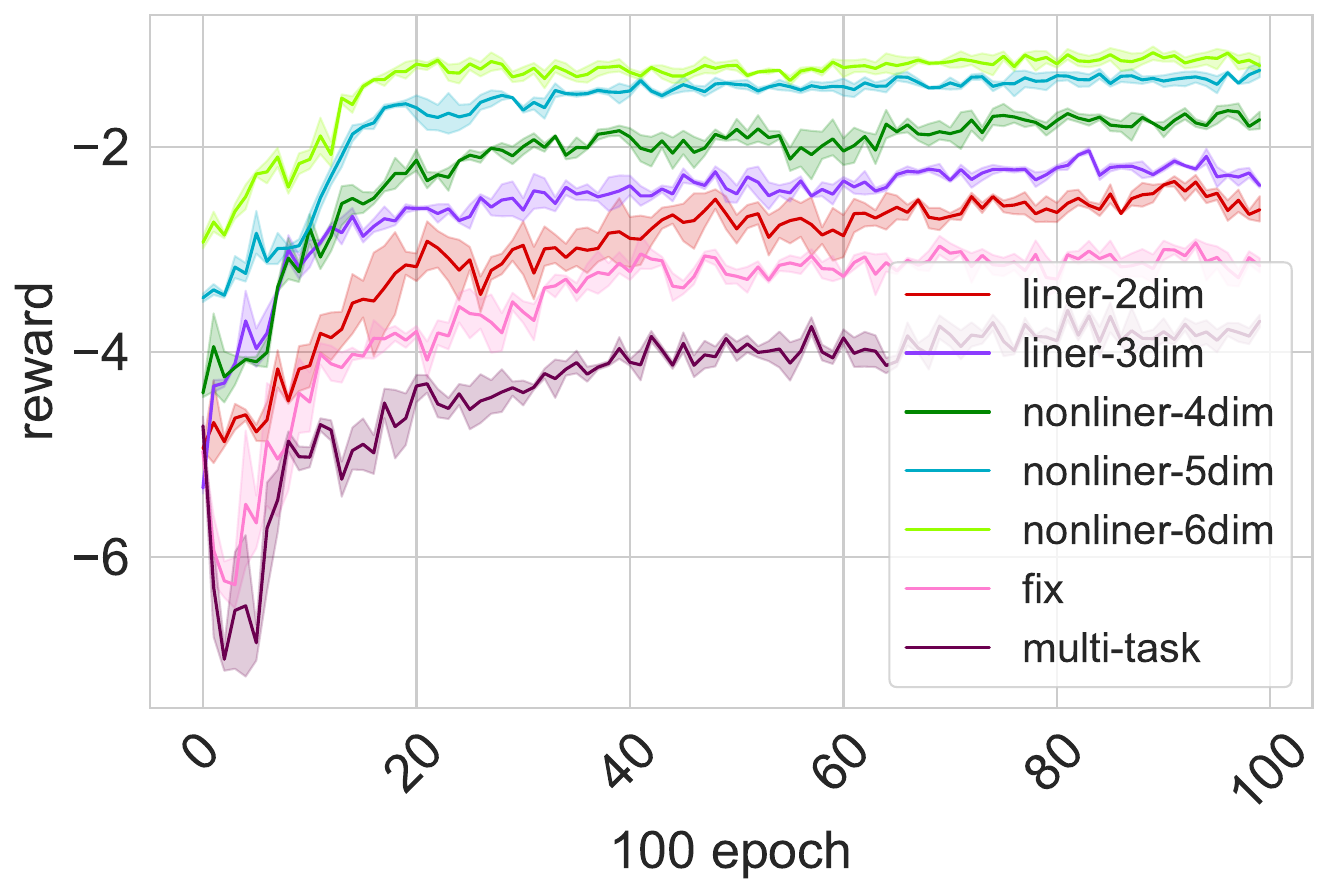}}\hfill
  \\
  \subfigure[N=5,M=2]{\includegraphics[width=0.33\textwidth]{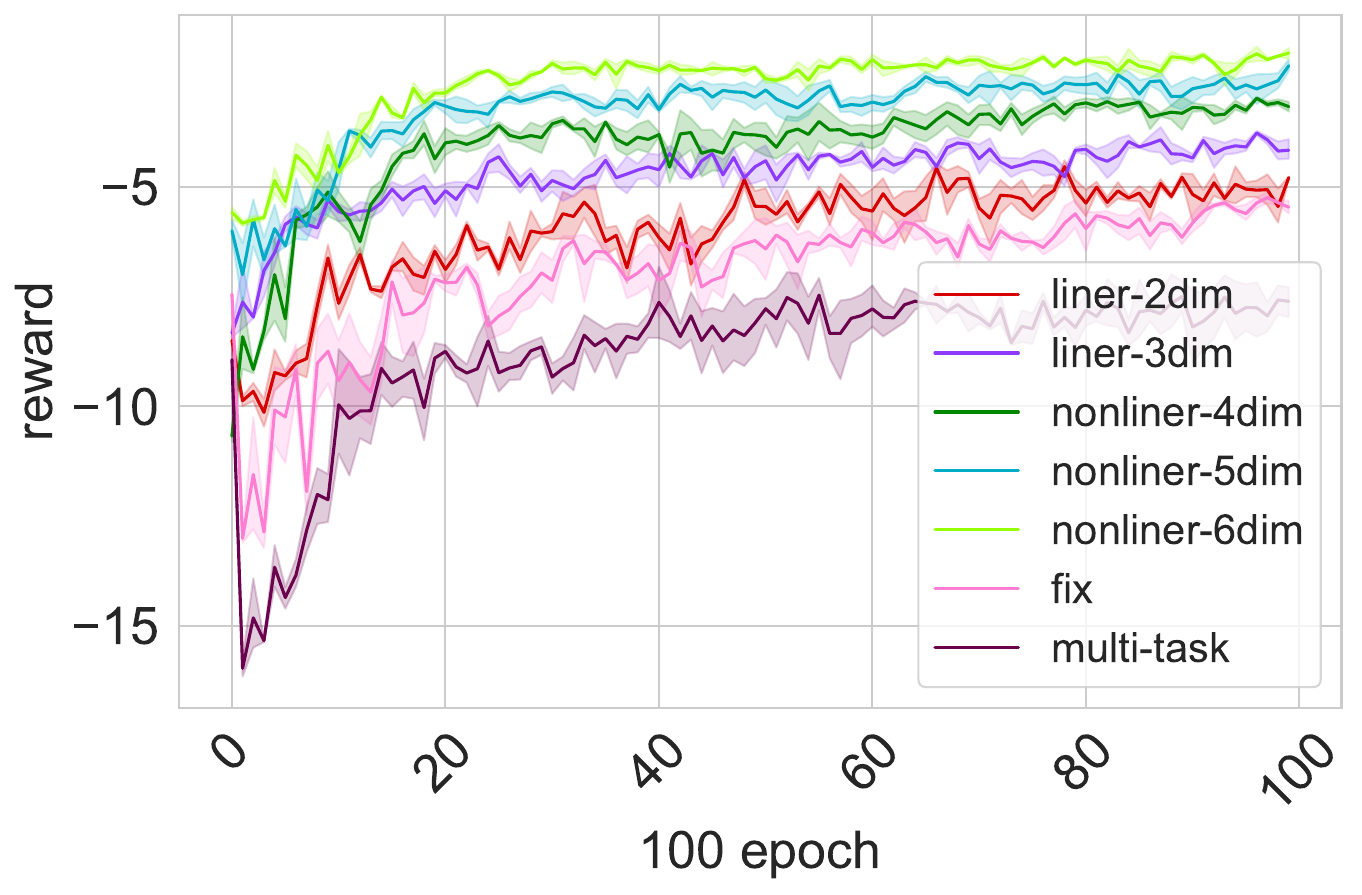}}\hfill
  \subfigure[N=5,M=4]{\includegraphics[width=0.33\textwidth]{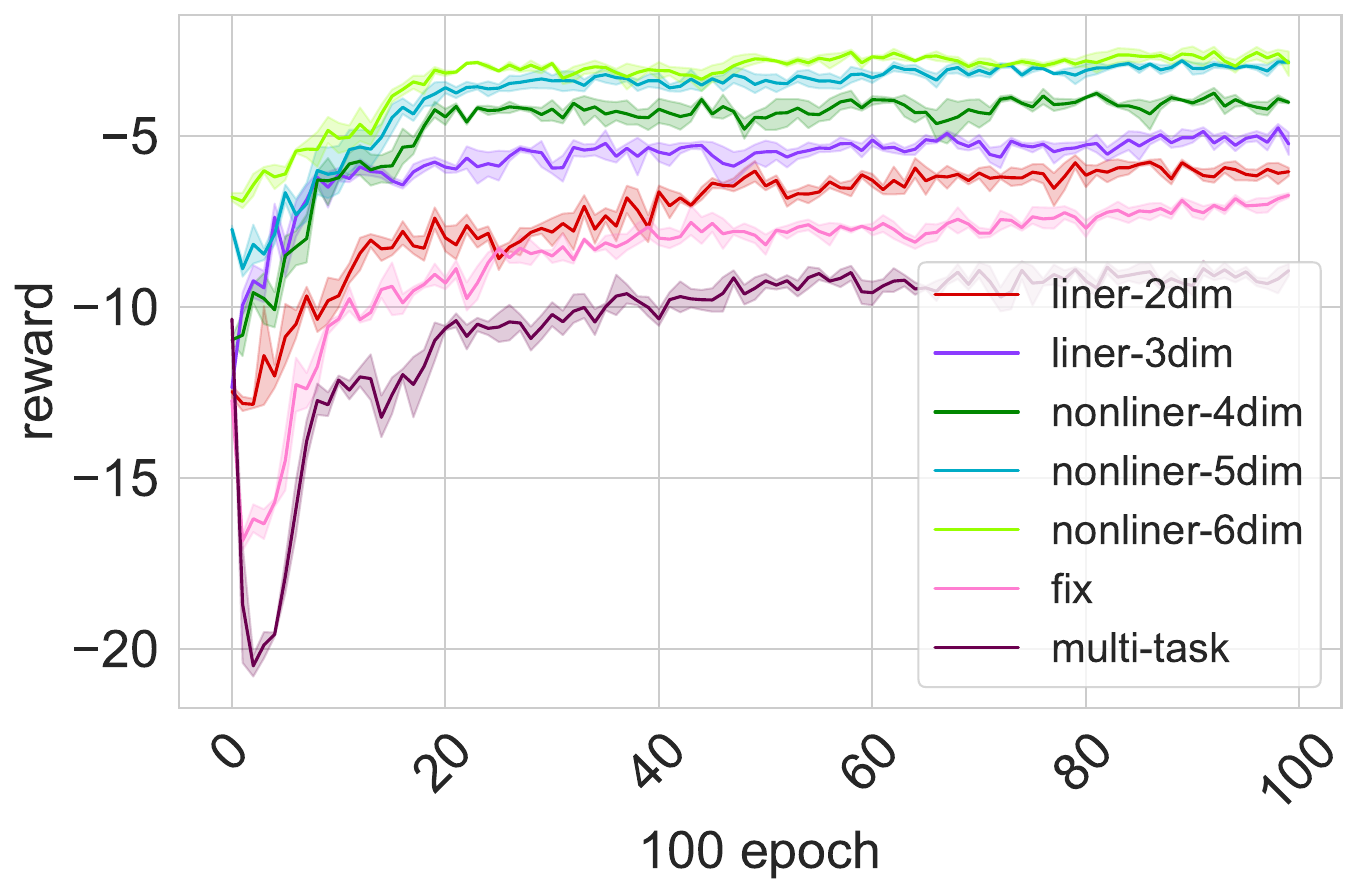}}\hfill
  \subfigure[N=5,M=7]{\includegraphics[width=0.33\textwidth]{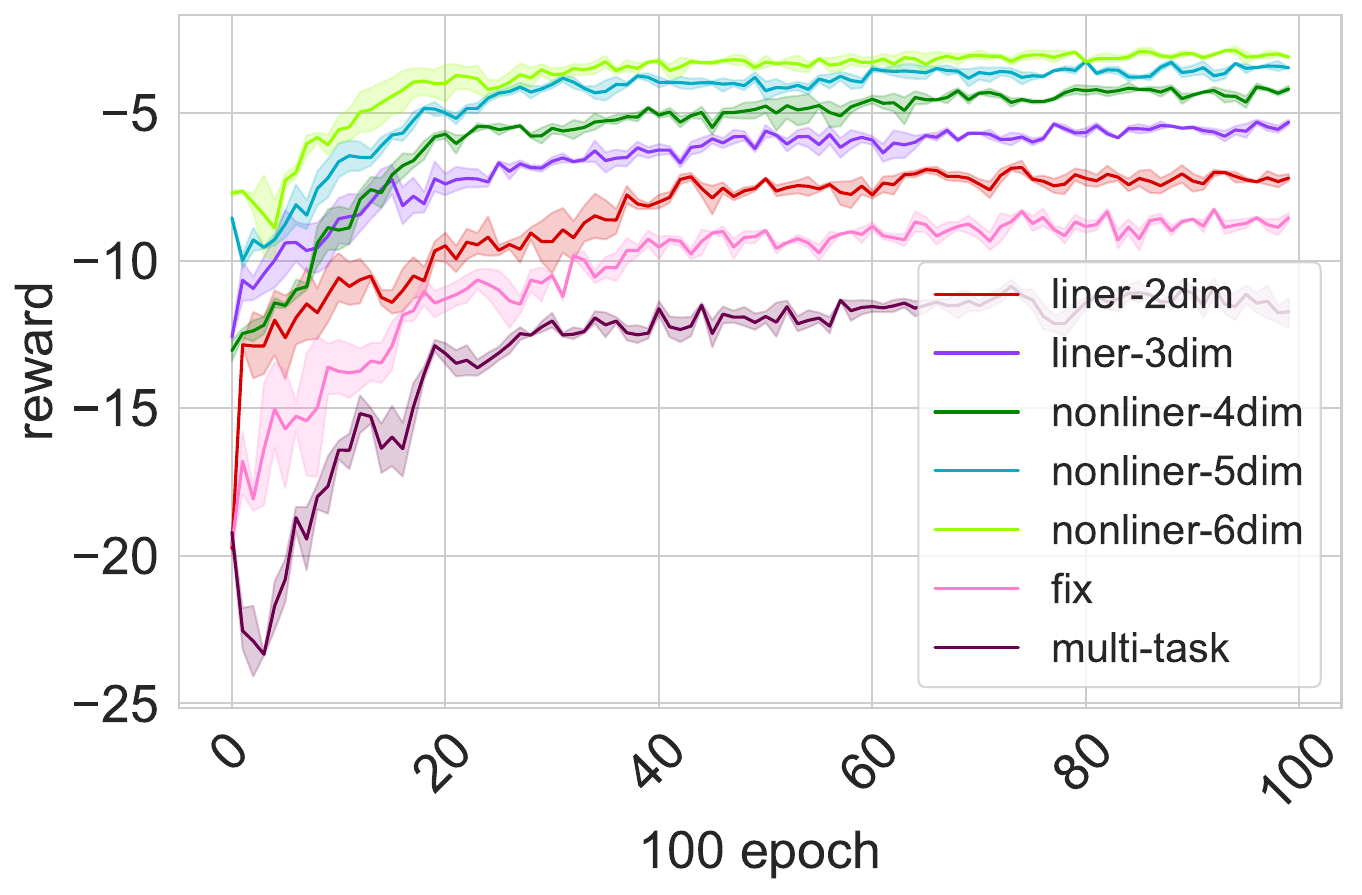}}\hfill
  \caption{Multiple sets of plots from the Ablation study. We can observe that our STUN pre-training framework achieved superior rewards in all cases.}
\end{figure*}

\subsection{SMAC}

\subsubsection{Implementation details and Hyper-parameters}
In this section, we introduce the implementation details and hyperparameters used in our experiments. For the environment, we removed the win rate as the reward and instead used the health of allied units and the damage inflicted on enemy units as the reward for training. This approach may lead to some convergence issues because a larger reward is always received at the beginning of the game. As the game progresses, the effect of the second reward gradually becomes apparent, meaning that these two types of rewards cannot reflect different latent styles in any particular situation. Moreover, the two rewards are not on the same scale, which can also affect the results. We still need to improve these issues in the environment. We used a set of hyperparameters for each environment, that is, we did not adjust hyperparameters for individual maps. Unless otherwise stated, we kept the same settings for the common hyperparameters shared by all algorithms, such as the learning rate, and kept their unique hyperparameters at their default settings.

Batch size $bs$ = 128, replay buffer size = 10000

Target network update interval: every 200 episodes

Learning rate $lr = 0.001$ 

td lambda $\lambda = 0.6$

Performance for each algorithm is evaluated for 100 (Enumerated all styles and took the average) episodes every 1000 training steps.

\subsection{Compare inverse learning with ground truth }
This is a supplement to the corresponding section of the main text, where we tested various map images. The results can be found in Fig.~\ref{fig:inverse_all}. The specific experimental setup is as follows: we tested using a Surrogate agent together with a STUN agent. We tested all of the Surrogate agent's $\mathcal{B}$ separately. We first fixed the Surrogate agent's $\mathcal{B}$ and performed tests with STUN using all of the $\hat{\mathcal{B}}$, where the trajectory size collected for the Surrogate agent was $n=300$ and for the STUN agent was $m=3000$. The values are $h = 0.03, h^{\prime} = 0.03$. KD-BIL requires us to define two types of distances, the first being the distance between reward functions $(dr)$ and the second being the distance between trajectories $(ds)$. For $(dr)$, if the reward function is linear, we use the cosine distance, and if it is nonlinear, we use the Euclidean distance. For $(ds)$, we use the Euclidean distance in all cases.
\begin{figure*}[h]
\label{fig:inverse_all}
\centering
  
  \subfigure[5m\_vs\_6m]{\includegraphics[width=0.15\textwidth]{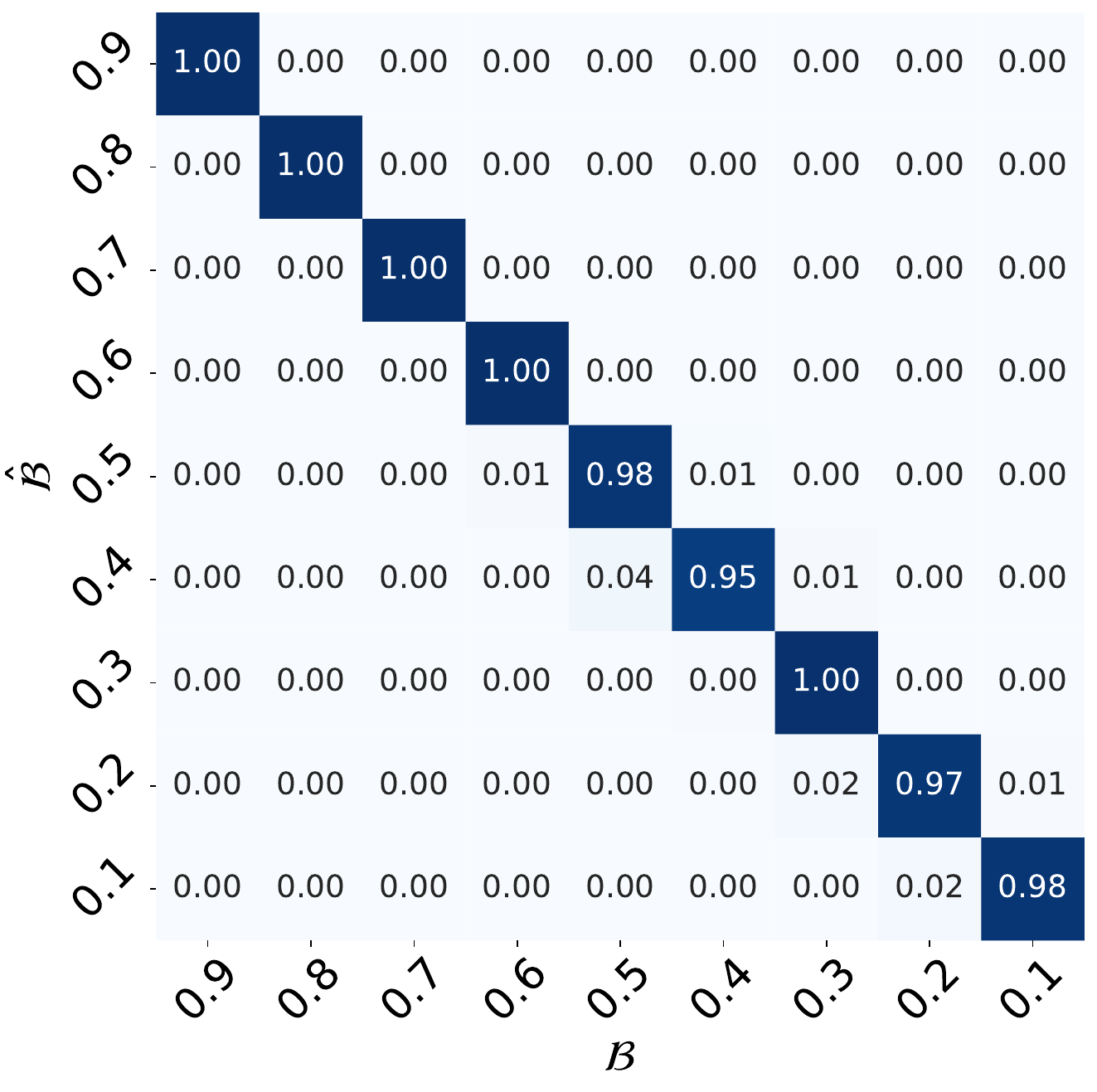}}\hfill
  \subfigure[3s\_vs\_5z]{\includegraphics[width=0.15\textwidth]{data/inverse/3s_vs_5z_mappo.pdf}}\hfill
  \subfigure[MMM2]{\includegraphics[width=0.15\textwidth]{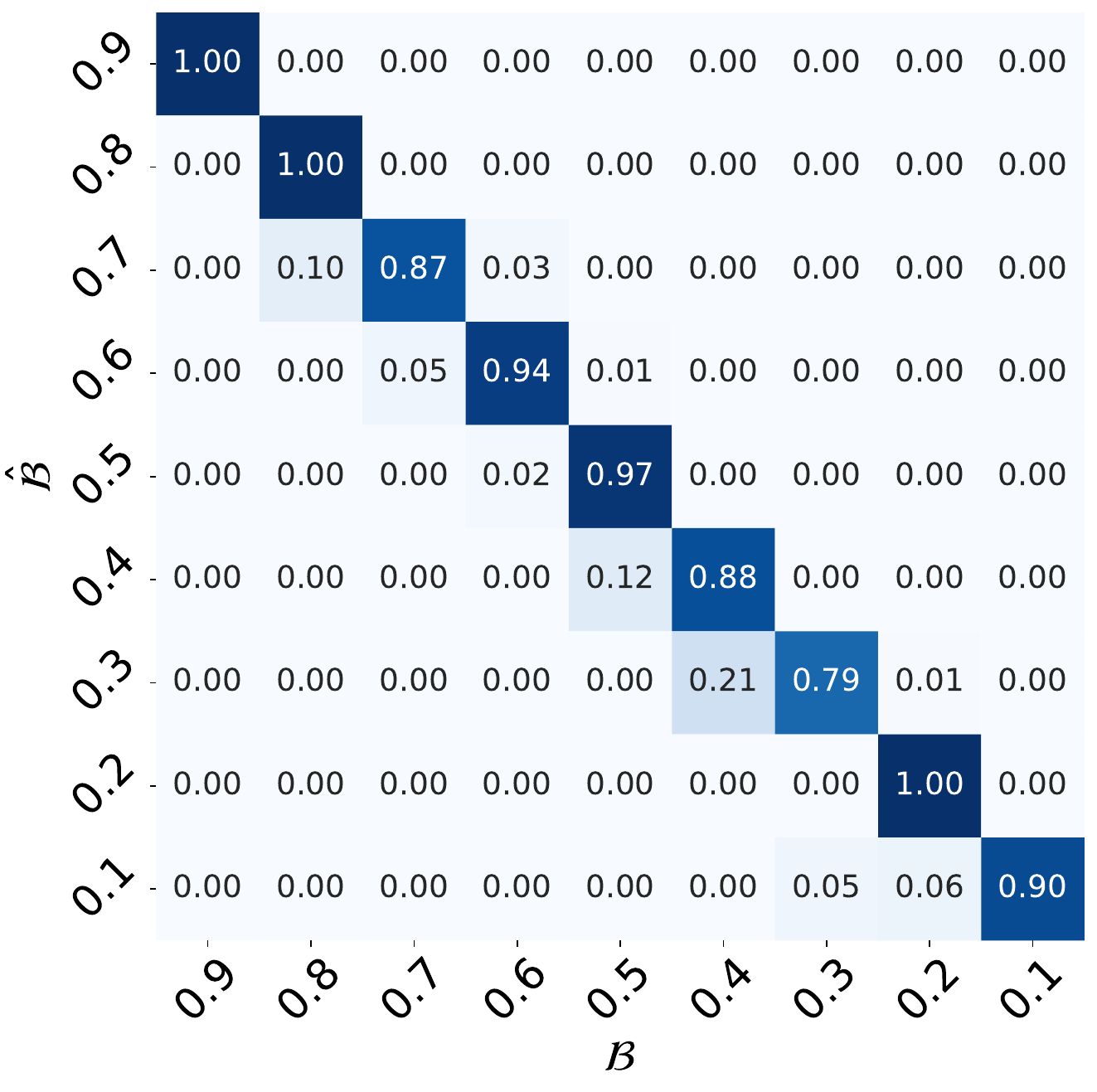}}\hfill
  \subfigure[27m\_vs\_30m]{\includegraphics[width=0.15\textwidth]{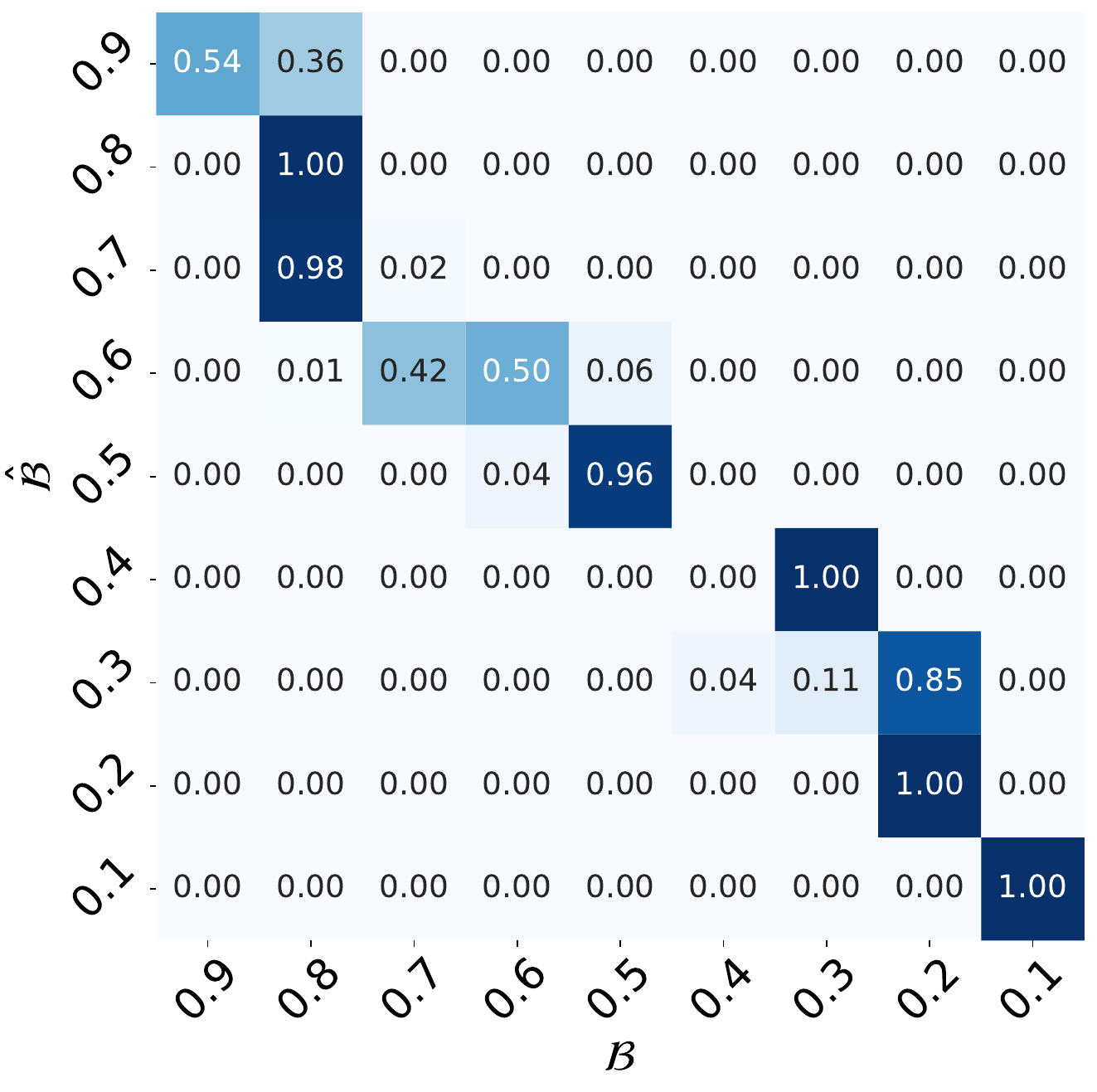}}\hfill
  \subfigure[6h\_vs\_8z]{\includegraphics[width=0.15\textwidth]{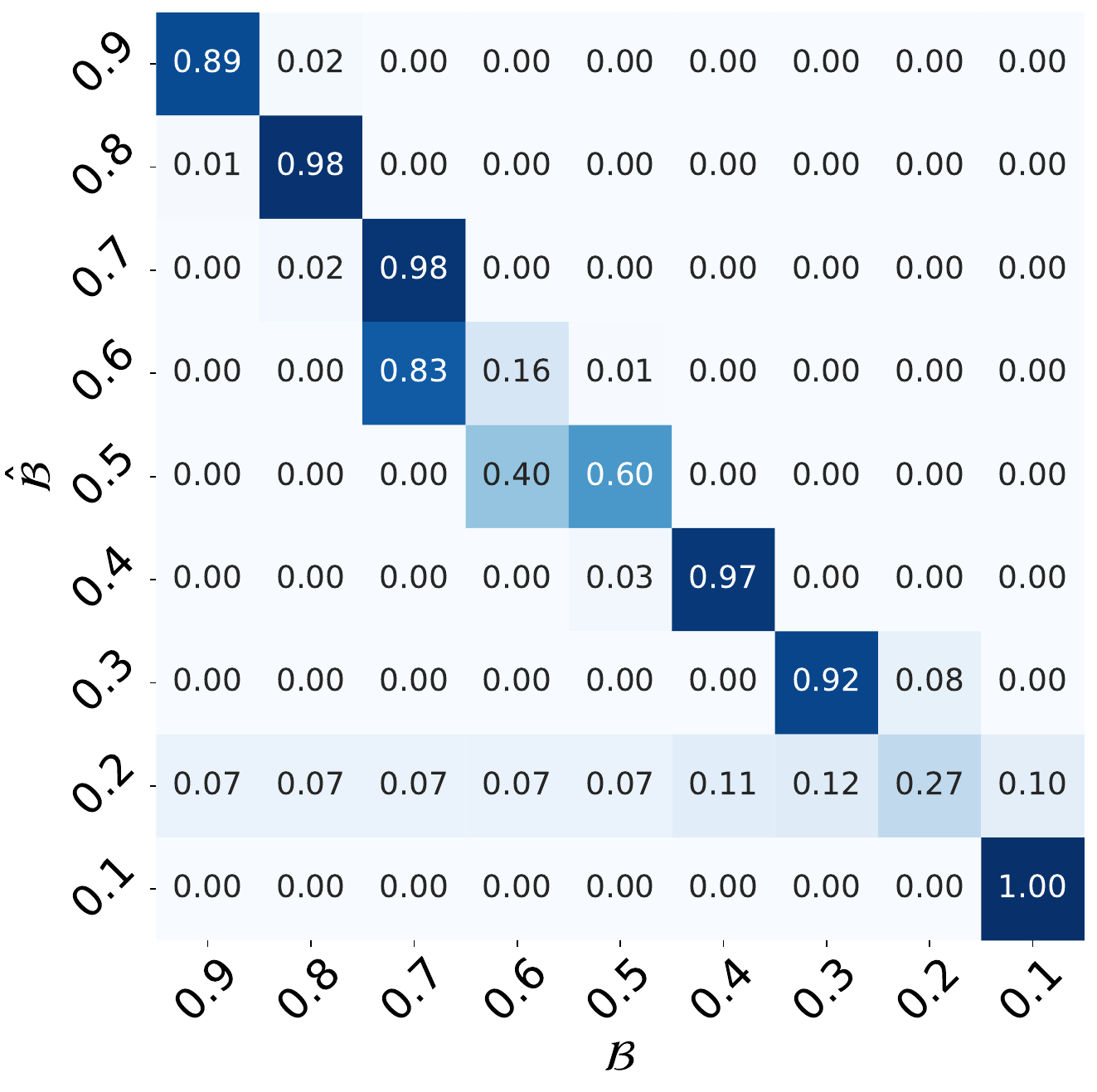}}\hfill
  \subfigure[corridor]{\includegraphics[width=0.15\textwidth]{data/inverse/corridor_mappo.pdf}}\hfill
  \caption{The estimation results of $\mathcal{B}$ obtained by KD-BIL in different map data. We can observe that in each map, the probability distribution of predicted $\hat{\mathcal B}$ are closely concentrated around ${\mathcal B}$, demonstrating the effectiveness of the prediction.}
\end{figure*}

\subsection{Compare reward on different map}
In this section, we primarily supplement the settings for Pre-training. During Pre-training, we sample $\mathcal{B}$ from a uniform distribution, with resampling occurring at each epoch. Additionally, we retest to obtain rewards under all styles every 100 epochs of training.

\subsection{Table comparing the performance with different unknown agents}
This section presents the results of running different unknown agents with different AI agents on other maps. We can observe that the STUN agent is always able to collaborate with different unknown agents to achieve superior results.
\begin{table*}[h!]
\label{tab:smac_5m_vs_6m}
\centering
\resizebox{0.985\textwidth}{!}{%
\begin{tabular}{@{}cclccclccccc@{}}
\toprule\toprule
\multirow{2}{*}{Unknown Agents} & Synergistic Agents &  & \multicolumn{3}{c}{Fixed-Behavior Agents} &  & \multicolumn{5}{c}{Multi-task Agents} \\ \cmidrule(l){2-12} 
           & \textbf{STUN} &  & FBA-C & FBA-B & FBA-A         &  & MAPPO & IPPO & COMA          & MAA2C & IA2C \\ \midrule
FBA-C     & \textbf{2.724} & & 2.558& 1.992 & 1.939& & 1.750& 1.962 & 1.313 & 0.784 & 1.071 \\
FBA-B & \textbf{3.490}& & 2.716& 3.377 & 3.264& & 2.547 & 2.735 & 2.320 & 1.132 & 0.679\\
FBA-A    & 6.290& & 4.475& 5.275 & \textbf{6.543} & &3.675& 4.928 & 3.373 & 1.554 & 1.456\\
\hline
MAPPO    & \textbf{3.792} & & 3.188& 2.754 & 2.981& &3.584& 2.849 & 1.396 & 1.566 & 1.339\\
IPPO & \textbf{3.584}& &3.452& 2.679 & 2.735& &2.924& 3.113 & 1.660 & 1.113 & 1.886\\
COMA    &\textbf{3.679} & &1.716& 2.886 & 2.622& &2.826& 2.641 & 2.735 & 1.622 &1.679\\
MAA2C    & \textbf{2.509} & &2.132& 2.320 & 2.018& &1.698& 2.264 & 1.113 & 2.398 & 1.037\\
IA2C    & \textbf{2.094} & & 1.924 & 1.377& 1.113 & &1.792& 1.962& 1.622 &  1.452 & 2.075\\
\hline
Average      & \textbf{3.520} &  & 2.770  & 2.833  & 2.902          &  & 2.600  & 2.807 & 1.942           & 1.453  & 1.403 \\
\bottomrule\bottomrule
\end{tabular}%
}
\caption{5m\_vs\_6m} 
\label{tab:my-table}
\end{table*}

\begin{table*}[h!]
\label{tab:smac_6h_vs_8z}
\centering
\resizebox{0.985\textwidth}{!}{%
\begin{tabular}{@{}cclccclccccc@{}}
\toprule\toprule
\multirow{2}{*}{Unknown Agents} & Synergistic Agents &  & \multicolumn{3}{c}{Fixed-Behavior Agents} &  & \multicolumn{5}{c}{Multi-task Agents} \\ \cmidrule(l){2-12} 
           & \textbf{STUN} &  & FBA-C & FBA-B & FBA-A         &  & MAPPO & IPPO & COMA          & MAA2C & IA2C \\ \midrule
FBA-C     & \textbf{1.551}& & 1.505& 1.383 & 1.454& &1.464 & 1.340 & 1.224 & 1.297 & 1.181 \\
FBA-B & 3.912 & & 3.486 & \textbf{4.128} & 3.777 & & 3.783 & 3.682 & 3.25 & 3.087 & 2.844\\
FBA-A    & 6.135 & & 5.967 & 5.945 & \textbf{6.340}& &5.470 & 5.621 & 5.551 & 4.605 & 4.437\\
\hline
MAPPO    & 4.168 & & 3.608& \textbf{4.668} & 4.094& &3.939& 3.844 & 3.128 & 2.885 & 3.094\\
IPPO & \textbf{4.479}& &3.689 & 3.777 & 4.060& &3.945& 3.757 & 3.006 & 3.101 & 2.925\\
COMA    & \textbf{3.959}& &3.541& 3.763 & 3.668& &3.5743& 3.081 & 3.682 & 2.898 & 2.256\\
MAA2C    & 3.75 & & 3.304& 3.655 & 3.608& & 3.398& 3.378 & 3.3108 & \textbf{3.837} & 3.033\\
IA2C    & \textbf{3.614}& &3.540& 3.479 & 3.421& &3.256& 3.304 & 2.939 & 3.252 & 3.565\\
\hline
Average      & \textbf{3.946} &  & 3.580  & 3.849  & 3.803          &  & 3.604 & 3.501 & 3.261         & 3.120  & 2.917 \\
\bottomrule\bottomrule
\end{tabular}%
}
\caption{6h\_vs\_8z} 
\label{tab:my-table}
\end{table*}

\begin{table*}[h!]
\label{tab:smac_27m_vs_30m}
\centering
\resizebox{0.985\textwidth}{!}{%
\begin{tabular}{@{}cclccclccccc@{}}
\toprule\toprule
\multirow{2}{*}{Unknown Agents} & Synergistic Agents &  & \multicolumn{3}{c}{Fixed-Behavior Agents} &  & \multicolumn{5}{c}{Multi-task Agents} \\ \cmidrule(l){2-12} 
           & \textbf{STUN} &  & FBA-C & FBA-B & FBA-A         &  & MAPPO & IPPO & COMA          & MAA2C & IA2C \\ \midrule
FBA-C     & \textbf{3.033} & &3.018 & 2.732 & 2.192& &1.398 & 1.128 & 1.141 & 1.063 & 1.024 \\
FBA-B & \textbf{6.129} & & 5.124 & 5.351 &  5.643& &3.675& 0.870 & 0.762 & 0.827 & 0.881\\
FBA-A    & \textbf{7.846} & & 6.683& 6.972 & 7.180& &5.913& 1.398 & 0.555 & 1.364 & 1.718\\
\hline
MAPPO     & \textbf{6.172} & & 5.052& 5.075& 5.886& &3.140 & 0.729 & 0.686 & 1.043 & 0.848\\
IPPO & \textbf{6.048} & & 4.572 & 5.470 & 5.654 & & 3.70& 0.924 & 0.762 & 0.902 & 0.859\\
COMA   & \textbf{6.216} & & 4.740& 5.189& 5.118 & &3.459& 0.675 & 2.697 & 0.982 & 0.913\\
MAA2C    & \textbf{5.529} & & 4.713 & 5.459 & 4.816 & & 2.670& 0.718 & 0.805 & 1.345 & 0.740\\
IA2C    & \textbf{5.410} & &5.362& 4.399 & 4.551& &2.800& 0.859 & 0.827 & 0.935 & 1.643\\
\hline
Average      & \textbf{5.798} &  & 4.908  & 5.081  & 5.130         &  & 3.344  & 0.913 & 1.029          & 1.058  & 0.986 \\
\bottomrule\bottomrule
\end{tabular}%
}
\caption{27m\_vs\_30m} 
\label{tab:my-table}
\end{table*}

\begin{table*}[h!]
\label{tab:smac_corridor}
\centering
\resizebox{0.985\textwidth}{!}{%
\begin{tabular}{@{}cclccclccccc@{}}
\toprule\toprule
\multirow{2}{*}{Unknown Agents} & Synergistic Agents &  & \multicolumn{3}{c}{Fixed-Behavior Agents} &  & \multicolumn{5}{c}{Multi-task Agents} \\ \cmidrule(l){2-12} 
           & \textbf{STUN} &  & FBA-C & FBA-B & FBA-A         &  & MAPPO & IPPO & COMA          & MAA2C & IA2C \\ \midrule
FBA-C     & \textbf{1.696} & & 1.582 &  1.376& 1.394& &0.725& 0.688 & 0.704 & 0.882 & 1.071 \\
FBA-B & \textbf{2.588}& &2.043& 2.376 & 1.820& &2.105& 2.067 & 1.354 & 0.923 & 1.768\\
FBA-A    &\textbf{3.423} & &2.347& 2.832 & 3.325& &2.473& 2.065 & 2.390 & 1.181 & 2.248\\
\hline
MAPPO    & 2.823& &1.607& 1.957&  2.066& & \textbf{3.029}& 2.271 & 1.523 & 1.255 & 1.695\\
IPPO & \textbf{2.490}& &1.521& 1.995 & 1.343& &1.932& 2.183 & 1.361 & 1.712 & 2.053\\
COMA    & \textbf{2.546}& &1.976& 1.285 & 1.436& &1.560& 2.141 & 2.410& 1.594 & 1.761\\
MAA2C    & \textbf{2.360} & & 1.723& 1.651 & 0.951& &1.882& 1.860 & 1.253 & 2.100 & 1.555\\
IA2C    & \textbf{2.288}& &1.760& 1.626 & 1.947& &1.975& 1.556 & 1.170 & 1.335 & 1.854\\
\hline
Average      & \textbf{2.527} &  & 1.820  & 1.887  & 1.785          &  & 1.960  & 1.854 & 1.521          & 1.373  & 1.751 \\
\bottomrule\bottomrule
\end{tabular}%
}
\caption{corridor} 
\label{tab:my-table}
\end{table*}

\begin{table*}[h!]
\label{tab:smac_MMM2}
\centering
\resizebox{0.985\textwidth}{!}{%
\begin{tabular}{@{}cclccclccccc@{}}
\toprule\toprule
\multirow{2}{*}{Unknown Agents} & Synergistic Agents &  & \multicolumn{3}{c}{Fixed-Behavior Agents} &  & \multicolumn{5}{c}{Multi-task Agents} \\ \cmidrule(l){2-12} 
           & \textbf{STUN} &  & FBA-C & FBA-B & FBA-A         &  & MAPPO & IPPO & COMA          & MAA2C & IA2C \\ \midrule
FBA-C     & \textbf{2.070}& &1.948& 1.702 & 1.653& &1.678& 1.396 & 0.941 & 0.998 & 0.989 \\
FBA-B & 3.955& &3.736& \textbf{4.332} & 3.833& &3.689& 2.480 & 0.959 & 1.094 & 1.236\\
FBA-A    & \textbf{7.507}& &4.846& 5.395 & 7.436& &4.507& 5.273 & 1.526 & 1.741 & 1.278\\
\hline
MAPPO    & \textbf{5.798} & &3.496& 3.852 & 4.189& &2.704& 2.822 & 1.274 & 1.145 & 1.050\\
IPPO & \textbf{4.273}& &3.586& 3.766 & 3.642& &3.201& 2.654 & 1.367 & 1.125 & 1.226\\
COMA    & \textbf{3.362}& &3.285& 2.156 & 2.512& &2.143& 1.305 & 2.341 & 1.433 & 1.015\\
MAA2C    & \textbf{3.161}& & 2.773 & 3.082 & 2.899& &2.220& 2.385 & 1.078 & 1.945 & 1.156\\
IA2C    & \textbf{3.245}& &2.248& 2.691 & 2.975& &2.046& 1.869 & 1.292 & 0.749 & 1.452\\
\hline
Average      & \textbf{4.171} &  & 3.240  & 3.372  & 3.642          &  & 2.774  & 2.523 & 1.347          & 1.279  & 1.175 \\
\bottomrule\bottomrule
\end{tabular}%
}
\caption{MMM2} 
\label{tab:my-table}
\end{table*}

\end{document}